\documentclass[graybox, envcountchap]{svmult}

\usepackage{makeidx}         
\usepackage{graphicx}        
\usepackage{multicol}        
\usepackage[bottom]{footmisc}

\usepackage{newtxtext}       %
\usepackage[varvw]{newtxmath}       

\usepackage[inline]{enumitem}

\usepackage{cite}
\usepackage[numbers,sort&compress]{natbib}
\usepackage[sectionbib]{chapterbib}

\usepackage{fix-cm}
\usepackage{amsmath,amssymb,amsfonts}
\usepackage{algorithmic}
\usepackage{booktabs}
\usepackage{subcaption}
\usepackage{tikz}
\usepackage{pgf-pie}
\usepackage{graphicx}
\usepackage[most]{tcolorbox}
\usepackage{xparse}
\usepackage{xspace}
\usepackage{enumitem}
\usepackage{textcomp}
\usepackage[dvipsnames]{xcolor}
\usepackage{float}
\usepackage{multirow}
\usepackage{tabularx}
\usepackage{lipsum}
\usepackage[rightcaption]{sidecap}
\tcbuselibrary{skins}
\usepackage{hyperref}

\makeatletter
\newcommand{\stitle}[1]{\smallskip\noindent\textbf{#1\@addpunct{}}}
\newcommand{\ititle}[1]{\smallskip\noindent\emph{#1\@addpunct{}}}
\makeatother

\definecolor{brown-web}{rgb}{0.65, 0.16, 0.16}
\definecolor{light-pink}{RGB}{251, 229, 229}

\newcommand{\red}[1]{\textcolor{red}{#1}}
\newcommand{\ex}[1]{\textcolor{brown-web}{#1}}

\usepackage[draft]{ifdraft}
\ifdraft{%
\usepackage[colorinlistoftodos, textwidth=3.2cm, loadshadowlibrary, shadow]{todonotes}%
\newcommand{\gilles}[1]{\todo[inline,color=blue!40,size=\scriptsize]{Gilles: #1}}
}
{
\usepackage[disable]{todonotes}%
\newcommand{\gilles}[1]{}
}

\definecolor{eastern-blue}{cmyk}{0.80, 0.13, 0.14, 0.04}
\definecolor{orient}{cmyk}{0.85, 0.44, 0.17, 0.14}
\definecolor{sea-green}{cmyk}{0.65, 0, 0.3, 0}

\newtcbox{\labelbox}[1][red]{
  on line, 
  arc=3pt, 
  colback=#1!10!white, 
  colframe=#1!50!black,
  boxrule=0.4pt, 
  boxsep=0pt,
  left=3pt, right=3pt, 
  top=0.5pt, bottom=0.5pt,
  before upper={\rule[-2pt]{0pt}{9pt}}, 
  fontupper=\small
}

\newcommand{\disease}[1]{\textbf{#1}\ \allowbreak\labelbox[BurntOrange]{Disease}}
\newcommand{\complication}[1]{\textbf{#1}\ \allowbreak\labelbox[Orchid]{\textit{Complication}}}
\newcommand{\anatomy}[1]{\textbf{#1}\ \allowbreak\labelbox[eastern-blue]{\textit{Anatomy}}}

\tcbset{
  pinkboxstyle/.style={
    colback=red!5!white,      
    colframe=FireBrick,       
    boxrule=1pt,              
    sharp corners,
  }
}

\newcommand{\eat}[1]{}
\usepackage[table]{xcolor} 
\usepackage{colortbl}      
\usepackage{lscape}

\makeatletter

\pretocmd{\@chapter}{%
  \if@mainmatter
    \pdfbookmark[1]{#1}{chapter.\thechapter}%
  \else
    \pdfbookmark[0]{#1}{chapter*.\thechapter}%
  \fi
}{}{}
\makeatother

\begin{document}

\frontmatter

\mainmatter

\title{Text Data Integration}
\titlerunning{Text Data Integration}

\author{Md Ataur Rahman\orcidID{0000-0002-9092-8771},\\
Dimitris Sacharidis\orcidID{0000-0001-5022-1483},\\ 
Oscar Romero\orcidID{0000-0001-6350-8328} and\\ 
Sergi Nadal\orcidID{0000-0002-8565-952X}
}

\institute{\textbf{Md Ataur Rahman} 
\at Universitat Polit\`{e}cnica de Catalunya, Barcelona, Spain, \email{md.ataur.rahman@upc.edu}
\at Universit\'e libre de Bruxelles, Brussels, Belgium, \email{md.ataur.rahman@ulb.be}
\and Dimitris Sacharidis \at Universit\'e libre de Bruxelles, Brussels, Belgium
\and Oscar Romero \and Sergi Nadal \at Universitat Polit\`{e}cnica de Catalunya, Barcelona, Spain
}
%
%


\maketitle 
\vspace{-10mm}

\abstract{
Data comes in many forms. From a shallow perspective, they can be viewed as being either in structured (e.g., as a relation, as key-value pairs) or unstructured (e.g., text, image) formats. So far, machines have been fairly good at processing and reasoning over structured data that follows a precise schema. However, the heterogeneity of data poses a significant challenge on how well diverse categories of data can be meaningfully stored and processed. \emph{Data Integration}, a crucial part of the data engineering pipeline, addresses this by combining disparate data sources and providing unified data access to end-users. 
Until now, most data integration systems have leaned on only combining structured data sources. Nevertheless, unstructured data (a.k.a. free text) also contains a plethora of knowledge waiting to be utilized. Thus, in this chapter, we firstly make the case for the integration of textual data, to later present its challenges, state of the art and open problems.
}
\keywords{text data integration, data discovery, data augmentation, data enrichment, natural language processing}

\vspace{-6mm}
\section{Introduction}
\vspace{-2mm}

In this chapter, we will motivate why there is a need to integrate textual data with structured sources. First, we discuss why integrating text remains challenging despite being the most widely available type of data. Next, we highlight the critical role that textual data can play in integration scenarios. In particular, we describe how textual data can mitigate data sparsity, enable data discovery, and enhance integration through data augmentation. Each section is accompanied by clear, motivating examples to concretely demonstrate the impact of integrating textual data.

\vspace{-6mm}
\subsection{Text Everywhere, Yet Hard to Integrate}
\vspace{-2mm}

Humans generate roughly 2.5 quintillion bytes of digitized data every day \cite{LVB18}. Apart from being in ``structured'' sources; these data can be found in numerous other ``unstructured'' formats such as web pages, weather forecasts, file servers, product reviews, scientific articles, commute maps, patient records, disease surveys, etc. Answering questions from a single source of data is relatively easy and suitable for small databases. However, a complex query we want to answer, could be requiring access over several heterogeneous data sources. Data integration plays a vital role in combining these distinct sources and providing the user with a unified view and query interface in such scenarios.

The task of data integration mainly falls into the domain of data management and engineering. But, when it comes to integrating data that are in unstructured formats such as texts, it is essential to incorporate ideas from several other domains such as the Semantic Web (SW), Machine Learning (ML), Natural Language Processing (NLP), and Knowledge Representation (KR) techniques. Again, each of the domains mentioned above solves a particular task in a different way. Hence, there is a need to develop a common framework for data integration techniques to benefit from all the aforementioned domains and yield solutions for heterogeneous data sources. On top of this, rather than following any schema, most state-of-the-art integration platforms that include textual sources store text data as plain instances \cite{ABC+22}. This leads to the same chicken-and-egg problem where some extra manual efforts are needed to extract structured information from those text instances before they can be utilized and integrated with other structured sources.

To aid this, a common way of representing disparate data sources is needed. As humans, we are good at inferring in such scenarios mainly because we perceive and represent information in our brains through conceptualization. \emph{Conceptualization} can be thought of as an abstract perspective of representing the knowledge we retain regarding the world around us \cite{Guizzardi06}. Every concept is expressed and linked in terms of articulated relations with some other concepts in this representation. Each concept, in turn, could be associated with its own real-world examples (e.g., attributes, causes and effects), and might also form hierarchical relations. Knowledge Graphs (KGs) are well known for their ability to store contextual information with data in a way that resembles the above scenarios. They can represent complex relationships between entities while maintaining semantic clarity through well-defined structures. This semantic richness, combined with their inherent capability to integrate heterogeneous data sources through common vocabularies, makes them particularly suitable for large-scale data integration tasks. Therefore, by creating a concrete and explicit manifestation of conceptualization for machines, KGs could lead to a better unification model for diversified sources of data.

\subsection{The Critical Role of Text in Data Integration}
\vspace{-2mm}

Unstructured data preserved in the form of natural language text is undoubtedly an invaluable source of knowledge. The integration of structured and unstructured data has become a critical area of research in recent years \cite{IC19}. Adaptation by the leading technological industries like Google, Microsoft, and Facebook emphasizes the significance of harvesting textual information with other structured sources in an enterprise. For example, Google integrates textual data from websites with structured databases (i.e., KGs) to enhance search results and provide richer answers in response to user queries. Similarly, Microsoft leverages textual data from emails and documents to improve their Office products and services, while Facebook combines textual content from posts with structured data to improve targeted advertisements and user recommendations. Integrating structured and unstructured data is important because structured data alone often captures only limited aspects of a topic, missing the contextual details that textual data naturally provides. Textual data can enrich structured datasets with domain knowledge, and implicit relationships that are typically absent in structured sources. Hence, integrating multi-modal data creates a more comprehensive view, enabling better understanding, accurate analysis, and sophisticated data-driven decision-making through improved inference over texts.
Manually converting all the heterogeneous data types into a single homogeneous structure involves numerous manual efforts and is often impossible, so we need to address the limitations of human intervention, ensuring adaptability to large-scale, real-world scenarios. Thus, the challenge is developing a scalable and automated data integration framework that can efficiently handle textual information alongside other structured sources. 

In the following, we justify the need of such a framework. 
Indeed, text could offer three key benefits when integrated with structured data: \emph{(i) mitigating data sparsity}, \emph{(ii) data discovery}, and \emph{(iii) data augmentation}.
We will discuss these scopes in the following sections, providing motivating examples to clarify their significance and impact.

\vspace{-4mm}
\subsubsection{Mitigating Data Sparsity}
\vspace{-2mm}
The data integration process often results in a large number of missing values (i.e., NULL), as each source provides only a partial view of the data. This issue, commonly referred to as the \emph{data sparsity problem}, arises when schemas across sources mismatch or the available structured data lacks sufficient context to fill in the missing values. Traditional value imputation techniques fail to address the heteroscedasticity (i.e., variance inconsistency) and the non-independent and identically distributed (Non-IID) nature of data integrated from different sources \cite{CYK22}. Moreover, these methods are limited to structured data sources and cannot effectively utilize unstructured textual data, which often contains valuable information for completing sparse datasets. This challenge can be mitigated by leveraging external textual sources to enrich integrated datasets through \emph{text conceptualization}.

\medskip

\stitle{Motivating Example-1:}
Consider the two structured datasets containing health-related information of Table \ref{tab:chapter1_structured_sources}. The first dataset, \textbf{D1}, provides information about diseases and their complications. The second dataset, \textbf{D2}, records diseases and their anatomical associations. When these datasets are integrated (Table \ref{tab:chapter1_integrated_table}), missing values ($\bot$) appear due to differences in their schema (Table \ref{tab:chapter1_integrated_table_a}). To address this sparsity, we turn to unstructured textual data, such as clinical texts. By integrating \textbf{conceptualized texts} with the structured data, we can enrich the integrated dataset with the missing values (Table \ref{tab:chapter1_integrated_table_b}). For instance, the missing \ex{\textit{Anatomy}} value for \ex{\textit{Tuberculosis}} can be filled using textual data describing its effect on the \ex{\textit{lungs}}. Similarly, the missing \ex{\textit{Complication}} values for \ex{\textit{Acoustic Neuroma}}, such as \ex{\textit{hearing loss}} and \ex{\textit{unsteadiness}}, can also be identified from the textual descriptions.

\vspace{-6mm}
\begin{table}[htbp]
    \centering
    \begin{subtable}[t]{0.40\textwidth}
        \centering
        \caption*{\textbf{D1}: Complication Dataset}
        \label{tab:chapter1_D1_table}
        \begin{tabular}{p{2cm}l}
            \toprule
            \textbf{Disease} & \textbf{Complication} \\ 
            \midrule
            Tuberculosis       & seizures \\
            Acne               & skin cancer \\
            \bottomrule
        \end{tabular}
    \end{subtable}
    \hfill
    \begin{subtable}[t]{0.5\textwidth}
        \centering
        \caption*{\textbf{D2}: Anatomy Dataset}
        \label{tab:chapter1_D2_table}
        \begin{tabular}{p{2.5cm}l}
            \toprule
            \textbf{Disease} & \textbf{Anatomy}  \\
            \midrule
            Acne                    & skin \\
            Acoustic Neuroma        & nervous system   \\
            \bottomrule
        \end{tabular}
    \end{subtable}
    \caption{Structured data sources containing joinable information.}
    \label{tab:chapter1_structured_sources}
\end{table}
\vspace{-6mm}

\begin{tcolorbox}[boxrule=0.5pt, boxsep=0pt, colback=light-pink]
\small{\textbf{\ex{Conceptualized Text:}} ``\disease{Acoustic neuroma} is a slow-growing non-cancerous \complication{brain tumor} effecting main (vestibular) \anatomy{nerve} leading from your \anatomy{inner ear} to your \anatomy{brain}. It can cause \complication{hearing loss}, and \complication{unsteadiness}. In contrast, \disease{Tuberculosis} generally damages the \anatomy{lungs}, leading to \complication{empyema}.''}
\end{tcolorbox}

\vspace{-6mm}
\begin{table}[htbp]
    \centering
        \begin{subtable}[t]{0.45\textwidth}
        \centering
        \caption{Integrated Data with Missing Values (\red{$\perp$})}
        \label{tab:chapter1_integrated_table_a}
            \begin{tabular}{ccc}
            \toprule
                \textbf{Disease} & \textbf{Anatomy} & \textbf{Complication} \\ \hline
                \rowcolor[HTML]{EFEFEF} 
                Tuberculosis & \red{$\perp$} & seizures \\
                Acne & skin & skin cancer \\
                \rowcolor[HTML]{EFEFEF}
                \begin{tabular}[c]{@{}c@{}}\rowcolor[HTML]{EFEFEF}
                Acoustic\\ \rowcolor[HTML]{EFEFEF}Neuroma\end{tabular} & \begin{tabular}[c]{@{}c@{}}\rowcolor[HTML]{EFEFEF}nervous\\ \rowcolor[HTML]{EFEFEF}system\end{tabular} & \red{$\perp$} \\
            \bottomrule
            \end{tabular}
        \end{subtable}
        \hfill
        \begin{subtable}[t]{0.45\textwidth}
        \centering
        \caption{Enriched Integrated Data}
        \label{tab:chapter1_integrated_table_b}
            \begin{tabular}{ccc}
            \toprule
                \textbf{Disease} & \textbf{Anatomy} & \textbf{Complication} \\ \hline
                \rowcolor[HTML]{EFEFEF} 
                Tuberculosis & \ex{lungs} & seizures \\
                Acne & skin & skin cancer \\
                \rowcolor[HTML]{EFEFEF}
                \begin{tabular}[c]{@{}c@{}}\rowcolor[HTML]{EFEFEF}
                Acoustic\\ \rowcolor[HTML]{EFEFEF}Neuroma\end{tabular} & \begin{tabular}[c]{@{}c@{}}\rowcolor[HTML]{EFEFEF}nervous\\ \rowcolor[HTML]{EFEFEF}system\end{tabular} & \ex{unsteadiness} \\
            \bottomrule
            \end{tabular}
        \end{subtable}
    \caption{Integrated data, (a) before and (b) after imputing missing values via text.}
    \label{tab:chapter1_integrated_table}
\end{table}
\vspace{-6mm}


\vspace{-4mm}
\subsubsection{Data Discovery}
\vspace{-2mm}
A common challenge in data integration is combining independently generated datasets that lack explicit connections or shared identifiers between them. These datasets, created in isolation from one another, often have no predetermined join keys or obvious relationships that would facilitate their integration. Data discovery through external textual sources introduces a new possibility for integrating such sources. By utilizing text, we could find conceptual links to produce join-paths across datasets. These conceptual links can be derived from textual sources discussing the same data elements. We can then identify implicit relationships between seemingly disparate data that traditional schema matching methods might miss. Thus, extending the integration scope beyond directly joinable tables.

\medskip

\begin{figure}[h]
    \centering
    \includegraphics[width=1.0\textwidth]{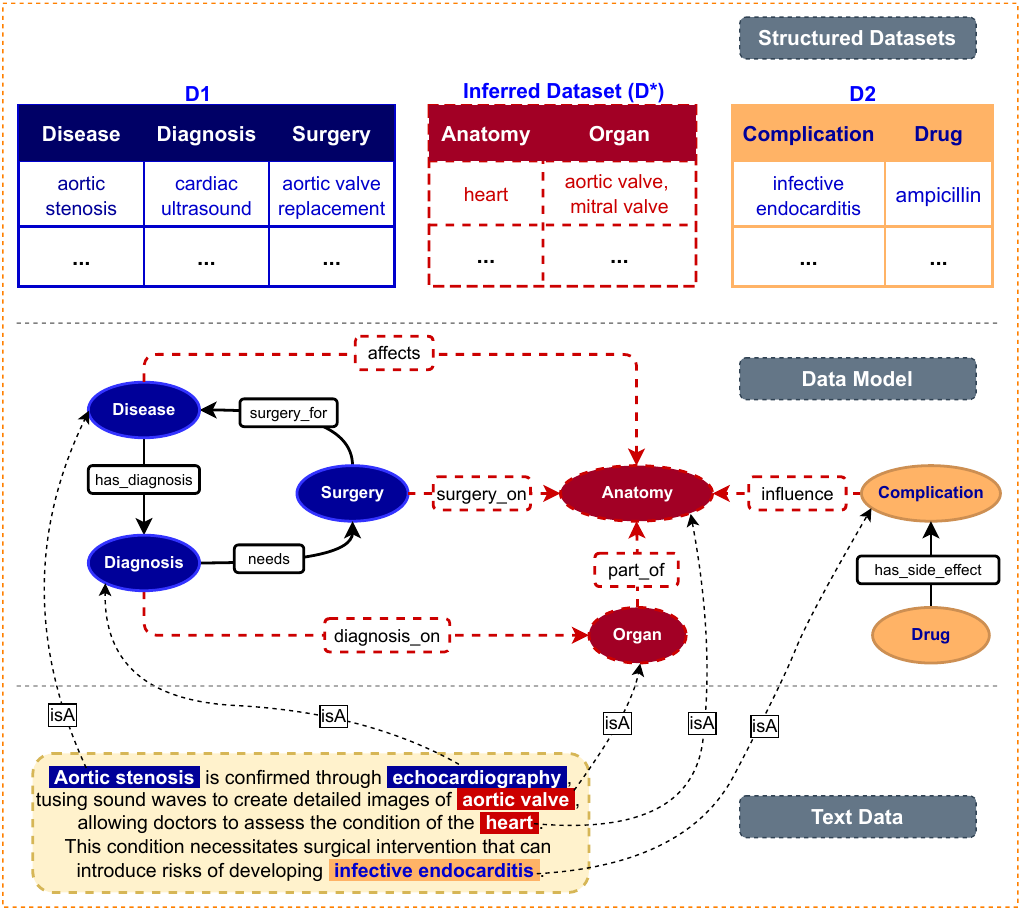}
    \vspace{-4mm}
    \caption{Discovering join-paths accross disjoint datasets via text.}
    \label{fig:chapter1_data_discovery_ex}
    \vspace{-4mm}
\end{figure}

\stitle{Motivating Example-2:} Consider the scenario in Figure \ref{fig:chapter1_data_discovery_ex}, where we start with two disjoint structured datasets \textbf{D1} (blue) and \textbf{D2} (yellow):

\begin{itemize}
    \item \textbf{Disease Dataset}: Contains diseases, diagnoses, and surgeries.
    \item \textbf{Complication Dataset}: Records complications and prescribed drugs.
\end{itemize}

\noindent These datasets lack explicit connections. However, textual data (bottom) from clinical books could help discover the inferred dataset \textbf{D*} (in red) along with new concepts and relationships (data model in the middle) such as:

\begin{itemize}
    \item \textbf{Anatomy}: Terms like \ex{\textit{heart}} are extracted as anatomical entities.
    \item \textbf{Organ}: Specific organs such as \ex{\textit{aortic valve}}.
    \item \textbf{Join-paths}: The identified join-paths via the inferred concepts/relations:
    \begin{center}
        \ex{\textit{Disease}} $\rightarrow$ \ex{\textit{Diagnosis}} $\rightarrow$ \ex{\textit{Organ}} $\rightarrow$ \ex{\textit{Anatomy}} $\rightarrow$ \ex{\textit{Complication}}
    \end{center}
\end{itemize}

\noindent Thus textual data could uncover join-paths via inferred concepts and relationships, enabling discovery and integration of disjoint datasets.

\vspace{-4mm}

\vspace{-8mm}
\subsubsection{Data Augmentation}
\vspace{-2mm}
Data augmentation is essential for data integration as it helps address the limitations of individual structured sources, such as incomplete schemas, sparse attributes, or missing relationships. By introducing additional concepts and relationships, augmentation ensures more comprehensive integration and enhances the overall utility of the combined data. This augmentation is crucial when combining two or more sources that do not contain any common attributes. For example, consider two tables from the same domain that could be integrated with a new relationship, or by simply adding a new column to one of the tables, leading to better integration. Textual data serves as a valuable source for identifying new concepts and relationships that can effectively augment and integrate structured datasets. This augmentation might involve extending the schema and/or enriching the instances of the structured datasets. In addition, schema evolution methods could enable the integration system to adapt to previously unknown information.

\vspace{-6mm}
\begin{table}[htbp]
    \centering
    \begin{subtable}[t]{0.40\textwidth}
        \centering
        \caption*{Patients Table}
        \label{tab:chapter1_patients_table}
        \begin{tabular}{llc}
            \toprule
            \textbf{PatientID} & \textbf{Name}       & \textbf{Age} \\ 
            \midrule
            P1                 & Alice Johnson       & 34           \\
            P2                 & Bob Smith           & 28           \\
            \bottomrule
        \end{tabular}
    \end{subtable}
    \hfill
    \begin{subtable}[t]{0.5\textwidth}
        \centering
        \caption*{Medications Table}
        \label{tab:chapter1_medications_table}
        \begin{tabular}{llr}
            \toprule
            \textbf{MedicationID} & \textbf{Drug}  & \textbf{Dosage} \\
            \midrule
            M1                    & Lisinopril     & 10mg           \\
            M2                    & Metformin      & 500mg          \\
            \bottomrule
        \end{tabular}
    \end{subtable}
    \caption{Source tables with disjoint structured data.}
    \label{tab:chapter1_source_tables}
\end{table}
\vspace{-6mm}

\vspace{-6mm}
\begin{table}[htbp]
    \centering
    \begin{subtable}[t]{0.40\textwidth}
        \centering
        \caption*{Associative Table: Prescription}
        \label{tab:chapter1_associative_table}
        \begin{tabular}{ll}
            \toprule
            \textbf{PatientID} & \textbf{MedicationID} \\
            \midrule
            P1                 & M1                    \\
            P2                 & M2                    \\
            \bottomrule
        \end{tabular}
    \end{subtable}
    \hfill
    \begin{subtable}[t]{0.5\textwidth}
        \centering
        \caption*{Integrated Data}
        \label{tab:chapter1_unified_view}
        \begin{tabular}{lclc}
            \toprule
            \textbf{Name}        & \textbf{Age} & \textbf{DrugName} & \textbf{Dosage} \\
            \midrule
            Alice Johnson        & 34          & Lisinopril       & 10mg           \\
            Bob Smith            & 28          & Metformin        & 500mg          \\
            \bottomrule
        \end{tabular}
    \end{subtable}
    \caption{Augmented tables showing integrated relationships.}
    \label{tab:chapter1_augmented_tables}
\end{table}
\vspace{-6mm}

\stitle{Motivating Example-3:} Consider the two tables from medical domain (in Table~\ref{tab:chapter1_source_tables}), where the \ex{\textit{Patients Table}} contains basic patient details such as \texttt{ID}, \texttt{Name}, and \texttt{Age}; whereas the \ex{\textit{Medications Table}} lists \texttt{Drug} and amount of \texttt{Dosage}. On their own, these two tables do not share any common attributes or connections. However, unstructured \textit{medical notes} could provide valuable relationships linking patients to medications. For example:

\begin{quote}
\textit{``Patient \ex{\textit{Alice Johnson}} presents with \ex{\textit{hypertension}}. Prescribed \ex{\textit{Lisinopril 10mg}} once daily. Advised to monitor blood pressure regularly.''}

\vspace{1mm}

\textit{``\ex{\textit{Bob Smith}}, diagnosed with \ex{\textit{type 2 diabetes mellitus}}. Started on \ex{\textit{Metformin 500mg}} twice daily. Recommended lifestyle modifications.''}
\end{quote}

\noindent By extracting these relationships from text, we can bridge the gap between the two tables and create an associative \ex{\textit{Prescription Table}} linking patients to their medications. This augmentation will allow queries to produce a unified view, such as retrieving patient information alongside their prescriptions (shown in Table~\ref{tab:chapter1_augmented_tables}). This demonstrates how textual data can lead to a comprehensive integration with through data augmentation.

\subsection{Challenges in Integrating Textual Data}
\vspace{-2mm}

Data sources vary in formats and applications. Structured and semi-structured data models such as relational data, JSON, or CSV may be useful for some use cases. Meanwhile, textual sources hold a wealth of information to combine with the former sources. Interpreting or querying over multiple such heterogeneous sources complicates the situation even more. Thus, integrating unstructured and structured data presents several challenges:

\begin{enumerate}
    \item \textbf{Heterogeneity of Data}: Structured data adheres to specific schemas, whereas unstructured data lacks such formalization. Extracting meaningful concepts and aligning them with existing schemas requires advanced processing techniques.
    
    \item \textbf{Semantic Ambiguity}: Textual data is inherently ambiguous, as words or phrases may carry different meanings depending on the context. Disambiguating these terms to identify their correct sense or underlying concept is a critical challenge.
    
    \item \textbf{Scalability}: Conceptualizing and integrating large volumes of textual data requires computationally efficient algorithms and representations.
    
    \item \textbf{Schema Evolution}: Static schema cannot accommodate new concepts and relationships extracted from texts. Evolving the schema dynamically to include new entities remains a critical open problem.
    
    \item \textbf{Knowledge Representation}: Representing extracted textual knowledge in a machine-understandable form that supports efficient inferencing and querying is essential yet challenging.
\end{enumerate}

\noindent To address these challenges, research in \emph{Information Extraction (IE)}, and \emph{KGs} has gained momentum, specially with the advent of \emph{Large Language Models (LLMs)}. KGs and LLMs, in particular, could provide a way of integrating both structured and unstructured data by allowing entities, concepts, and relationships to be represented in an interconnected graph format. We will discuss these possibilities in detail throughout Section \ref{sec:chapter1_SOTA-ESR-1.1}.

\vspace{-6mm}
\subsection{Objectives}
\vspace{-2mm}

This chapter presents a comprehensive overview of the field of \textbf{Text and Structured Data Integration}, focusing on the following key aspects:

\begin{enumerate}
    \item \textbf{Understanding the Challenges}: Identifying the fundamental challenges in integrating structured and unstructured data, including heterogeneity, ambiguity, schema evolution, and knowledge representation.
    \item \textbf{Key Techniques and Approaches}: Highlighting state-of-the-art methods in data integration, information extraction, and ontology learning that enable the conceptualization of textual data.
    \item \textbf{Existing Solutions, Gaps and Open Problems}: Identifying prominent research contributions and the gaps that remain unresolved, particularly in text conceptualization and information extraction domains.
\end{enumerate}

\noindent Rather than proposing a specific framework, the focus remains on presenting existing literature, highlighting its contributions, and identifying future research directions. Additionally, we aim to provide a thorough understanding of the challenges, approaches, and advancements related to structured and unstructured data integration.


\vspace{-6mm}
\section{State of the Art}
\label{sec:chapter1_SOTA-ESR-1.1}
\vspace{-2mm}

\begin{figure}[htb]
  \centering
     \centering
     \includegraphics[width = 0.5\linewidth]{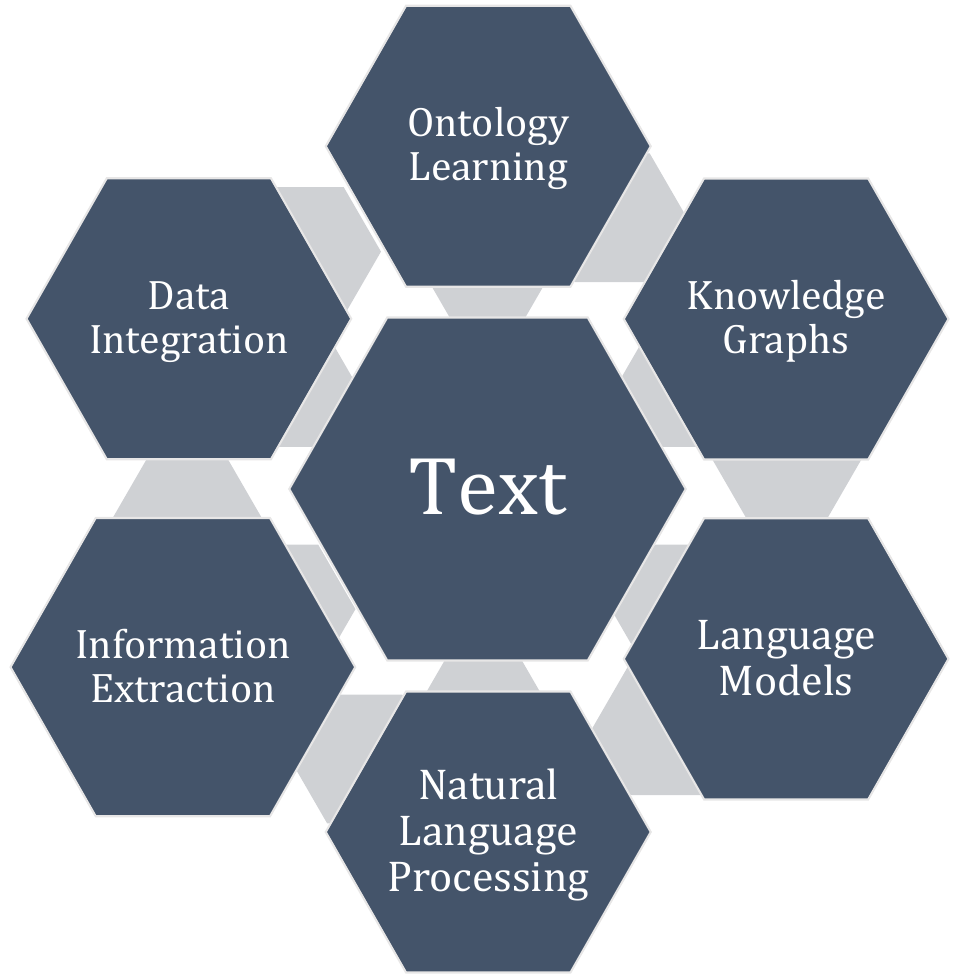}\
     \vspace{-1mm}
     \caption{Relevant Fields for Data Integration with Text.}
     \label{fig:chapter1_related_fields}
     \vspace{-2mm}
\end{figure}

In order to achieve data integration that considers natural language text with structured data, we have to combine methods and techniques from several disciplines such as KGs, LLMs, NLP, IR, DI, and Ontology Learning (OL). Text integration requires solutions from all these fields, since they are not enough on their own to solve the task. Thus, we will arrange this section in a way that reflects the relationship of textual integration with the most prominent literature related to these domains (shown in Figure \ref{fig:chapter1_related_fields}).

\vspace{-2mm}

\vspace{-6mm}
\subsection{Data Integration}
\label{sec:chapter1_sota-data-integration}
\vspace{-2mm}

Traditional approaches to data integration were predominantly focused on \texttt{ETL} (Extract, Transform and Load) processes \cite{NHP+20, DRPH22} for ingesting and cleaning the data before loading them into a data lake or data warehouse \cite{AP12}. However, one of the pitfalls of these classical approaches to data integration is - dealing with the data \emph{variety} challenge at a large scale. In solving the variety issue, one must assume that, integration must be done over a multitude of data formats (i.e., text, xml, csv, relational) coming from structured, semi-structured and unstructured sources.

The integration of multiple data sources can be either based on a \textit{physical} or a \textit{virtual} approach. While the former consolidates the original sources, the latter however, keeps the sources intact through modeling an integrated view (a.k.a. mediated schema) \cite{WP03}. Thus, individual local schemas of the associated data sources are mapped together to form a single virtual global schema. The modeling of different sources to the integrated view could be achieved via either Global-As-View (GAV) \cite{CGH+94, RS97} or Local-As-View (LAV) mappings \cite{LRO96}. While the constructs of the global schema are represented as views over the local schemas in the GAV-based approach, the opposite applies to the LAV mapping. However, a better choice is to combine both approaches, which is commonly referred to as Both-As-View (BAV) \cite{MP03} or Global-and-Local-As-View (GLAV) \cite{FLMO99} mapping. As the data sources adhere to different data models, part of the integration process involves transforming them into a common data model. This is typically done through a Hypergraph-based Data Model (HDM) \cite{PM98}. One such implementation of HDM uses KGs by utilizing Resource Description Framework (RDF) triples \cite{MMMO04, APR12}. In this way, each source schema is mapped into a local RDF schema which sometimes is referred to as the \textit{local graph}. Eventually, the local graphs are used to form the integrated \textit{global graph} (which is also an RDF named graph) through the use of a (semi-) automatic global schema builder \cite{TRJ15, JNR+21}.

Discrepancies in schemas and vocabularies used by different data sources while modeling the same domain of interest are termed as conceptual or \emph{\textbf{semantic heterogeneity}} \cite{Wang17}. Resolving this type of heterogeneity is considered as one of the major hurdles in data integration \cite{GHMT17}. In dealing particularly with unstructured data such as texts, traditional fixed schema-based integration techniques are incapable of providing a satisfactory result while resolving the semantic heterogeneity, mainly due to the schemaless nature of that data \cite{DNR+09, AAKA20}. For this reason, a popular way is to first construct an intermediate form of structure or schema out of text sources \cite{BCF+08}. This task is formally known as \textit{Ontology Learning} (OL) \cite{GM04} and nowadays often interchangeably referred to as \textit{Knowledge Graph Construction} (KGC) from the text.

\vspace{-6mm}
\subsection{Ontology Learning and Knowledge Graph Construction}
\vspace{-2mm}

An \emph{ontology} can be described as an `explicit specification of conceptualization' \cite{Gruber93} that encodes domain-specific implicit knowledge by utilizing some form of a semantic structure. In layman's terms, we may view an ontology as a conceptualization of a particular domain. The terms OL \cite{MS00b} and KGC are often used in the same context based on their underlying knowledge representation principles. While these two terms are seen to be denoting the same formalism, there is a vague but subtle difference between them. Ontologies are highly structured knowledge that facilitates queries and reasoning over complex relationships and can be used to infer new knowledge. They are fundamentally designed to capture very complex relationships between classes and individuals. On the other hand, KGs are often used to represent or instantiate ontologies in their simplest form without worrying too much about the integrity constraints and formal semantics. Thus, by combining the power of semantic web technologies with NLP and IE \cite{MHL20}, ontologies in the form of KGs can be (semi-) automatically constructed from textual data \cite{KI18, PS18, MLR18, HOS+24} in order to extract and represent knowledge contained within the text \cite{BCM05, BC08}.

\begin{figure}[h]
  \centering
     \centering
     \includegraphics[width = 0.7\linewidth]{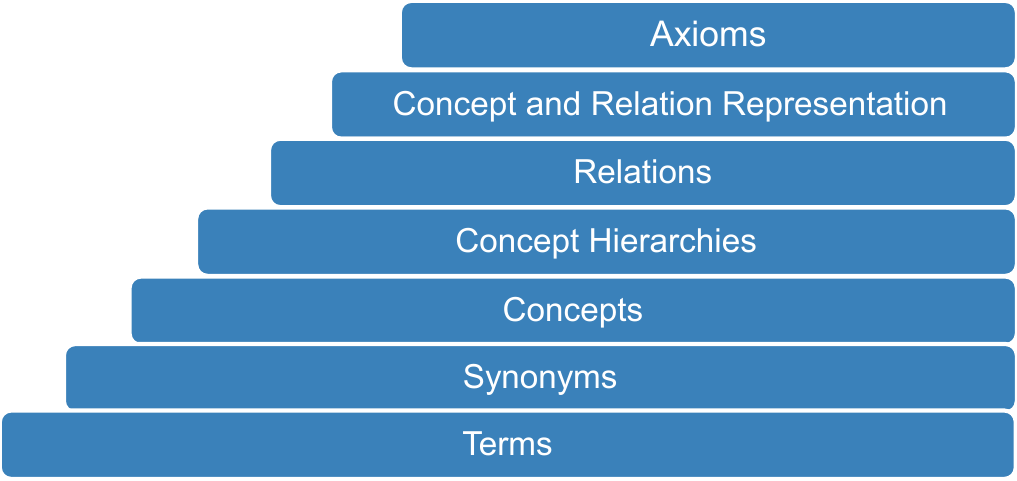}
     \caption{Extended Layer Cake for OL, KGC and KR from Text (extended from \cite{BCM05}).}
     \label{fig:chapter1_layer-cake}
     \vspace{-4mm}
\end{figure}

Figure \ref{fig:chapter1_layer-cake} depicts an extended framework that includes \textit{Concept and Relation Representation} (originally proposed by Buitelaar, 2005~\cite{BCM05}) for different layers of outputs that are commonly targeted (partially or fully based on the task and domain at hand) while constructing KGs from textual sources. In the following, we correlate the tasks and approaches to achieve these layers of outputs with the help of the most prevalent practices within this domain.

\medskip

\stitle{Term:} A term can be defined as a word or phrase from the natural language text, having a meaning associated with it. Term extraction is the initial requirement for KGC. Terms are the individual entities that act as a basis for identifying concepts and relations. Extracting terms from textual sources mainly uses NLP techniques. Approaches such as \textit{Named Entity Recognition (NER)}, \textit{sentence parsing}, \textit{part-of-speech (POS) tagging} \cite{Abney97} and \textit{morphological analysis} are commonly used in identifying the terms of interest that are associated with a domain \cite{SB04}. Again, the above can be achieved via either purely rule-based or following statistical and probabilistic ways \cite{KAG21}. Recent advances \cite{Ruder20} in NLP and ML techniques also pushed the boundaries of the aforementioned approaches.

\medskip

\stitle{Synonyms:} Although linguistically synonyms are words or tokens that denote the same meaning, in OL, it is mostly a phase where terms that are associated with similar concepts are clustered together. While terms can be seen as purely syntactic properties of language, synonyms are associated with the semantics of those terms. Earlier work for this intent uses synsets such as WordNet \cite{Miller95}, or EuroWordNet \cite{Vossen98} which links words through semantic relations in order to represent synonyms. However, as there could be different meanings associated with a particular term depending on the context and domain at hand, it is important to determine the relevant sense. \textit{Word-sense disambiguation} \cite{WWF20} tries to resolve this issue in general. However, for OL and KGC, it is important to extract and disambiguate domain-specific sense of a term \cite{MS00a, BS02, TPT+00}. Other techniques from information retrieval and text mining domain such as \textit{heuristic} \cite{LP01} and \textit{conceptual clustering} \cite{LP02} in parallel with \textit{latent semantic indexing} (such as LSA, LSI) \cite{LD97} and \textit{topic modeling} (such as LDA) algorithms \cite{BNJ03} are shown to be effective in domain-specific semantic disambiguation. A more recent approach within the NLP community, however, follows the use of \textit{word embedding} techniques such as Word2Vec \cite{MSC+13}, FastText \cite{BGJM17} or GloVe \cite{PSM14} in order to solve the \textit{semantic textual similarity} (STS) task. STS can help determine the word, phrase as well as sentence-level semantic similarities more effectively, even for a very specific domain \cite{HRJM15}.

\medskip
    
\stitle{Concepts:} The notion of concept is a highly disputed philosophical topic.\footnote{\url{https://plato.stanford.edu/entries/concepts/}} In purely linguistic terms, a concept is anything that we conceive in our mind when we hear or read something. It is a mental representation of entities that exist in our brain. Although the literature \cite{ACG20a, BCM05, WLB12} in the context of our discussion, profiles concepts as purely being entities formed by grouping or clustering semantically similar terms (i.e., synonyms) \cite{LP02, GEM20}, we do not distinguish concepts from entities, or attributes at this stage. The idea is to identify and extract concepts from the text that will, later on, be categorized as being either one of the two. For the purpose of data integration, the concepts might come from the global graph of already integrated structured sources. This also narrows down the domain for the specification of what might be the viable concepts. The most basic approach to concept extraction in the form of entities involves the use of NER and \textit{co-reference resolution} \cite{EN12, ETE17}. Another way is to use \textit{syntactic parsing} techniques to identify domain-specific noun phrases (NPs) as concepts \cite{GGA+02, RCOK13}. Entity and relations could also be jointly extracted from text using \textit{universal dependency parsing} where subject-predicate-object triples are identified \cite{LXZ+08, EN12}. However, a more recent practice to such concept extraction utilizes neural \textit{Language Models} (LMs) based on the \textit{transformer architecture} along with \textit{attention mechanisms} (e.g., BERT, XLNet, RoBERTa, T5) in a supervised way~\cite{HLS+22, MT22}.

\medskip
    
\stitle{Concept Hierarchies:} The process of obtaining a more general, higher-level concept from a set of lower-level concepts can be defined as the formation of concept hierarchy \cite{HKP11}. For instance, \textit{Vegetables} and \textit{Fruits} could be combined to form a higher-level concept, namely \textit{Food}. It is often termed as \textit{taxonomic relations}, which defines the hierarchical relationships among concepts. Lexico-syntactic patterns \cite{Hearst92, BOS04} are one of the most popular ways of obtaining such hierarchies in the form of hypernym/hyponym relationships. It uses syntactic and grammatical rules within the same sentence, such as `$<$NP$>$, $<$NP$>$ and/or other $<$NP$>$' (e.g., `Oranges, Apples and other Fruits') to classify the taxonomic relationship. A different paradigm uses \textit{distributional semantics} or \textit{co-occurrence analysis} with the assumption that ``a word is characterized by the company it keeps" \cite{Harris54, Firth57, Sahlgren08}. Word embedding is originally inspired by this distributional hypothesis and can be utilized to build semantic hierarchies of concepts given a large corpus to learn from \cite{FGQ+14, NKWV17, AMB19}.

\medskip

\begin{figure}[h]
  \centering
     \centering
     \includegraphics[width = 0.9\linewidth]{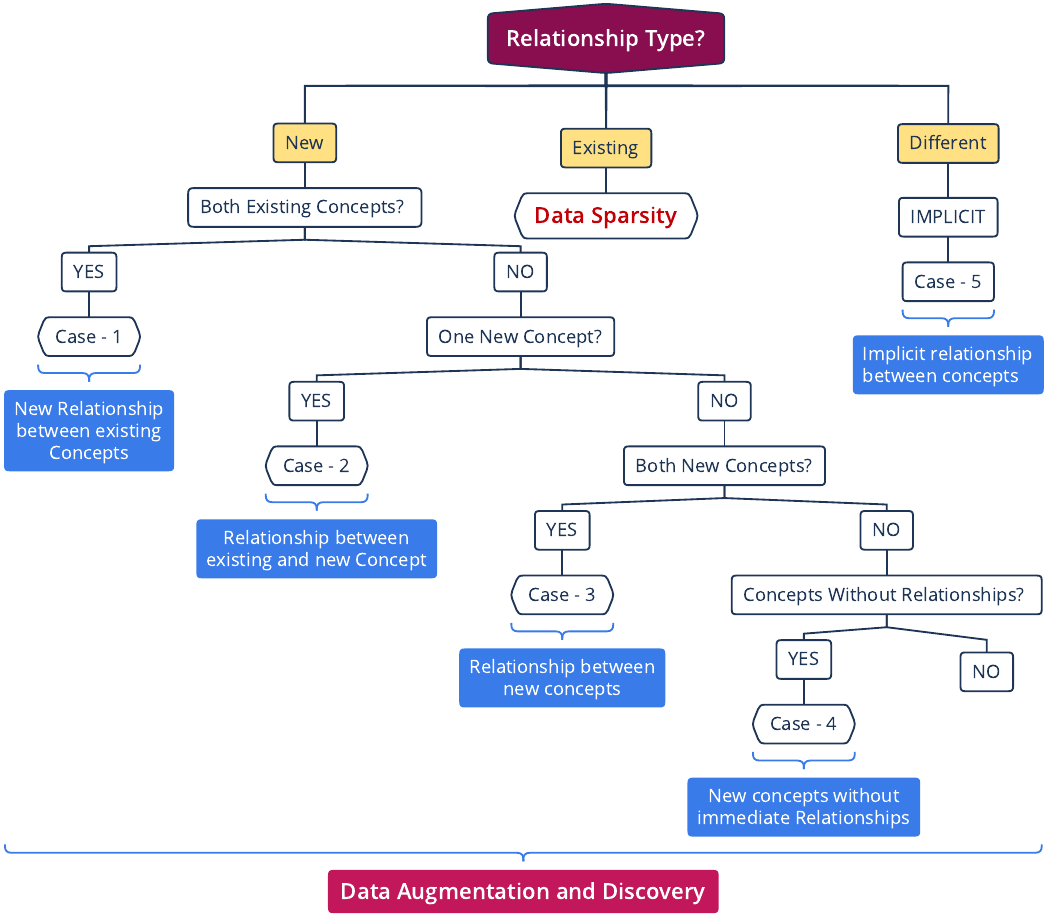}
     \caption{Types of Relationships depending on the underlying Concepts.}
     \label{fig:chapter1_concept_rel_types}
\end{figure}

\stitle{Relations:} The goal here is to identify non-taxonomic relationships between the concepts discovered in earlier stages. Non-hierarchical textual relations are mostly denoted by attributes, thematic/semantic roles (i.e., agent, patient, location, source, or goal), meronymy (part-of relation), causality (e.g., smoking causes lung cancer), and possession (i.e., my, your, his, her, have) \cite{PW13}. As mentioned earlier, the task of concept (entities) and relation extraction can be done separately or jointly \cite{ACG20b} and is considered a research area itself. There are many ways to discover relations from the text \cite{AMS18}. Extracted potential relations might take the form of either \textit{binary} ($n=2$) or \textit{n-ary} ($n\geq3$) types. Subject-verb-object \textit{triple extraction} \cite{AR19, LXZ+08} is an example of a binary relation extraction task. Traditional unsupervised approaches \cite{HGY+19, EN12} to such triple extraction often make use of syntactic and semantic parsing along with some feature engineering such as POS tagging. Some of the other ways are - \textit{association rule mining} using seed \textit{knowledge base} (KB) \cite{MS00b}, \textit{dependency structure analysis} \cite{QCL+18}, \textit{thematic roles extraction} for relation phrase association, \textit{semantic role labeling}, \textit{discourse-based} relation extraction, \textit{frame parsing}, domain-specific verb pattern matching \cite{KC18}, and utilizing \textit{semantic templates} as seeds \cite{YWGH14}. On the other hand, \textit{distant supervision} techniques \cite{MBSJ09, MMM+16} and LMs \cite{NTW11, HGY+19, HLS+22} are shown to be effective in this area as well. For conceptualizing text, the schema at hand (e.g., global graph) could either contain this concept or could be a completely new one. Based on these criteria, relationships might either have zero, one, or two new concepts associated with it. Depending on the type of extracted concepts, we can categorize relationship types differently (as shown in Figure \ref{fig:chapter1_concept_rel_types}).

\medskip

\stitle{Concept and Relation Representation:} As of now, semantic web, and data management technologies have made huge progress in data storage, representation, and retrieval. Yet, they failed to provide a system capable of learning and producing high-level representations, the sort of concepts humans exploit during language usage and comprehension. Humans are capable of using those higher-order concepts in order to generalize and reason in robust ways. It is something even an infant could do while the machine struggles with it. KGs are grounded on the principle of providing a layer of abstraction to data through graph-based configurations. A recent trend \cite{JPC+22, TAG21} shows that KGs are being extensively deployed in scenarios that require integrating and extracting information from multiple, possibly diverse data sources at a large scale. Apart from this, it is easier to manage and \textit{query} over KGs, making it a perfect candidate while representing text for data integration purposes \cite{LSD+21}. However, the most notable works in this field try to embody texts in the form of a \textit{data graph} (instances represented by nodes and edges) and do not follow any schema while doing so \cite{CILC19, MYH+19, ABC+22, SPBS22}. This approach of keeping text \textit{instances} as it is \cite{MDD22}, fails to facilitate reasoning over them. Instead, a more reasonable approach should be to model data as a KG having dynamic \textit{schemata}, the intention of which is to specify a high-level hierarchical structure with semantics \cite{RCM+23}. Thus, an evolving semantic schema could aid in specifying the meaning of high-level terms (a.k.a. terminology or vocabulary) manifested by the KG, which will eventually facilitate reasoning over the data \cite{HBC+21}. This might give KGs the supremacy we sought for modeling human-like high-level representation of abstract concepts and reasoning, the kind that we use in language understanding or while solving complex problems. For this purpose, representation languages such as RDF, RDFS, and OWL could be used.

\stitle{Axioms:} Axioms are rules and constraints that assert and govern the interaction between concepts and relations by adding expressivity into the ontology \cite{BM15}. Yielding a set of rules and axioms that explicitly captures knowledge in a particular domain is the foremost aim of OL \cite{WLB12}. Frequently expressed by first-order logic or description logics, axioms are the top-level output in the OL process (see Figure~\ref{fig:chapter1_layer-cake}). The task can be divided into \textit{discovering axioms} and \textit{learning axioms}. Automatic derivation of such rules from text is still a challenging and less explored area \cite{GS15, BM15, KAG21}. Classic approaches to extract axioms include \textit{inductive logic programming} \cite{FV11, KB15} or the usage of \textit{axiom templates} in the form of linguistic patterns \cite{SMD12}.

\vspace{-6mm}
\subsection{Information Extraction}
\label{sec:chapter1_sota-information-extraction}
\vspace{-2mm}

One of the biggest challenges of integrating structured and unstructured data is extracting the information from the unstructured sources into a common structure. The problem can be approached from a textual information extraction perspective by conceptualizing unstructured text documents using the concepts present in the integrated schema. This could also aid the data sparsity problem for data generated from an integration system. Data integration focuses on providing a unified view of data over a set of disparate and heterogeneous sources \cite{Lenzerini02}, and typically combines the underlying datasets with operators that allow for partial matches, such as \textit{outer join}~\cite{LP76} or \textit{full disjunction}~\cite{RU96}. 
The consequence, however, is the generation of a large number of missing values (a.k.a.\ \textit{labeled nulls}), which are reported to account for 15\% of the values~\cite{SI18}, since each data source captures a different set of instances (particularly, if these datasets have been independently generated). Further, each data source provides a partial view of the data of interest, thus even after combining them the resulting integrated data presents an incomplete view. This is commonly referred to in the literature as the \textit{data sparsity problem}~\cite{AGG06, XQX+15}.

Missing value imputation is a very common data cleaning technique in the data integration pipeline to enhance data quality~\cite{IC19}. Focusing on repairing structured data, such cleaning approaches can be categorized into constraint-based and learning-based ones. Systems in the former category (e.g., LLUNATIC~\cite{GMPS20}, NADEEF~\cite{DEE+13}, or HORIZON~\cite{ROA+21}), aim to model and enforce data dependencies or business rules modeled as denial constraints~\cite{Fan15}. Alternatively, those in the latter (e.g., HoloClean~\cite{RCIR17}, Baran~\cite{MA20}, or Garf~\cite{PST+22}), automatically generate data repairs leveraging pre-trained probabilistic models or models trained from the available structured data. Despite their effectiveness, the approaches above are limited to the available structured sources to learn and compute repairs. A significant portion of data organizations hold resides in texts~\cite{CIKW16},  which is largely unused. The idea of enhancing data cleaning methods with external information has already been adopted in systems such as KATARA~\cite{CMI+15}, using knowledge bases and crowdsourcing, or Cleenex~\cite{PFLG24}, that leverages user feedback to define an iterative process. Yet, to the best of our knowledge no missing value imputation solution exists that exploits raw textual data (i.e., without annotations). On the one hand, the data management community has proposed several information extraction techniques~\cite{FKRV16}, aiming to extract relational views from textual sources. However, the generation of structured views from textual data is limited to those extraction rules that can be defined over the text~\cite{FKRV15}. On the other hand, NLP community has proposed text conceptualization approaches~\cite{Sarawagi08} which aim to automatically label text span with a predefined label.

State of the art techniques for text conceptualization include (a) transformer-based LMs to perform entity recognition from text, and (b) zero-shot approaches with LLMs. However, the former require domain-specific large annotated data with rich context~\cite{SZM+22}. At the same time, they also need to be re-annotated and re-trained when the reference schema changes, while they also suffer from bias towards the most frequent entity, inconsistent performance, and demand vast resources to be trained. 
In such a data integration setting, LMs perform poorly when trained with structured data only~\cite{KBD22}. The latter approaches, despite their massive popularity, often fail due to their uncertainty and tendency to hallucinate~\cite{CZZ23, ASM23}. LLMs struggle to maintain recall and precision when prompted with large text and complex concept categories~\cite{ZZG+24}. Because of their attention mechanism, they often overlook fine-grained instances in the text, which are crucial for IE. Furthermore, resource-wise, it is not feasible to train or fine-tune an LLM like GPT-4 on a case-by-case basis, a situation that often pertains to organizations typically dealing with their own data for integration purposes~\cite{YHD+22}.

Recent advances in data science have enabled novel data integration techniques that extract and integrate information from diverse data sources on a large scale \cite{JNR+21}. Despite being a well-researched area in data engineering, the task of mitigating incomplete data during data integration is often overlooked due to its inherent challenges \cite{YK20}. To tackle this problem, we can apply existing techniques to extract, organize and enrich \cite{AC15} structured information from unstructured or semi-structured data. One such strategy is \textit{Entity-Centric Slot-Filling}~\cite{PPF+21, GRG21} where incomplete information is extracted via conceptualization with regard to the concepts defined within the data integration system.

\medskip

\stitle{Statistical Data Imputation:} Most conventional techniques \cite{Pigott01} to impute data rely on statistical measures such as mean substitution, frequent category (mode) imputation, and maximum likelihood estimates; they are predominantly focused on either categorical, numeric, or quantitative data \cite{KH20, JAB21}. Although most techniques perform quite well for numeric data \cite{JPR19, BDA21}, their performance decreases for qualitative data \cite{BSS+18}. From a statistical point of view, data imputation in data integration scenarios is more challenging than data imputation in single-database scenarios \cite{YK20}, since traditional statistical techniques do not deal with the \textit{heteroscedasticity} (i.e., variance inconsistency) and \textit{non-independent and identically distributed} (i.e., Non-IID) nature of data integrated from different sources \cite{CYK22}. As a consequence, a common practice is to simply ignore missing values \cite{RWS+23}. Methods that try to mitigate qualitative missing values are either too limited in the sense that they only consider a fixed number of variables \cite{RS04} or heavily dependent on domain-specific large datasets \cite{BSS+18}.

\medskip

\stitle{Entity Recognition:} The problem of extracting information from textual data has been deeply studied in the IE and NLP communities. IE techniques encompass several tasks, being one of them \emph{entity recognition} (ER) \cite{FWRR17} \cite{EN12}, which intends to identify entities mentioned in natural language text into pre-defined categories. Yet, traditional ER techniques involve creating lexicons or dictionaries containing a list of entities with specific tagging rules \cite{QMR+16}. Although these approaches are useful on small domains, they often tend to fall short in scenarios having complex entity types. Developing those rules also requires strong domain knowledge; hence, these systems are not generic. Alternatively, machine learning (ML) methods, such as Support Vector Machines (SVM), Decision Trees, Hidden Markov Models (HMM) \cite{Zhao04}, and Conditional Random Fields (CRFs) \cite{LSS08, RWL12, LWL15}, have also been employed for this task. These are trained in a supervised fashion using annotated text samples with entity labels and ad-hoc feature engineering. However, since these models solely rely on training examples, they struggle to generalize to new, unseen data. 

Most advanced state-of-the-art ER techniques use (L)LMs based on transformer architectures with attention mechanisms (e.g., BERT, RoBERTa, T5, XLNet, GPT, LlaMA, UniversalNER). LMs exploit contextual information (i.e., neighboring words) and distributional semantics (words with similar distributions have similar meanings) \cite{LS23} via an unsupervised pre-training phase. Then, they apply a supervised strategy called \textit{masked language modeling} (MLM) \cite{DCLT19, LOG+19, AMB+19}.
These techniques, in the presence of a large corpus of annotated text, outperform previous ER techniques~\cite{BDM23, WJB+21, HLS+22, MT22, WSC+22, ZZG+24}. Regardless of the technique used, we find two main approaches when predicting the entity label in an ER technique. Either identify the label by only considering the text at hand or the text and a reference schema. The former is a common practice in the knowledge base population and enrichment~\cite{LLX+20, JG11, VGN+21, SWC+21, XZX+23, TST23}, where they extract information from text in the form of \textit{subject-verb-object} relations and use entity linking~\cite{RMD13} techniques to enrich the KB with the extracted subjects and objects~\cite{APM15, CILC19, MYH+19, ABC+22, SPBS22, AKB+22}. These techniques rely on syntactic categories (e.g., noun phrases) and tend to generate long-tail entities that require further processing before being used~\cite{MDD22}. For this reason, the current trend is to conceptualize entities in the textual data following a reference schema~\cite{RCM+23}. State-of-the-art techniques~\cite{Wilks97, MHL20, DDL+24} can be used in this regard and convert the information embedded in the text into a structured representation via conceptualization. This is also the chosen approach to facilitate the enhancement~\cite{XZX+23, FRE+20} and construction~\cite{JM25, KI18, PS18, MLR18, FEF+23} of structured knowledge sources. \textit{Slot-Filling}~\cite{ZZC+17, GRCG21, ZC21} is another well-known technique that utilizes ER in order to reduce data sparsity via conceptualized text~\cite{RNRS24}. Slot filling involves populating entity-specific templates (e.g., \textit{[Acoustic Neuroma, causes, $<$slot$>$]}) with information extracted from text. It can be either \textit{document-centric}, focusing on entities represented by a single document, or \textit{entity-centric}, where information about a concept is spread across multiple documents in a corpus. By leveraging external text sources, missing information can be filled in, enhancing the quality and usefulness of the integrated data.

\vspace{-6mm}
\subsection{Advanced Language Modeling: Going Beyond}
\label{sec:chapter1_sadvanced_llms}
\vspace{-2mm}

Previous attempts at automatically extracting concepts have proven challenging since they almost always relied on a fixed set of concept categories (i.e., static schema)~\cite{FBB+23}, making it impossible to build KGs from raw text without knowing the associated concepts in advance~\cite{ACG20a}. With the cutting-edge advancement of (Large) Language Models, it is now possible to materialize this task. We examine these approaches in detail below: 

\medskip

\stitle{Contextual Embeddings:} Contextual embeddings such as ELMo and BERT \cite{DCLT19} have changed how words are represented based on surrounding texts. These methods often perform better than static embeddings because they take into account polysemy and contextual details, which makes concept modeling more accurate across a wide range of domains \cite{TGG20}. Techniques like pooled contextualized embeddings \cite{ABV19} achieved state-of-the-art ER results, while domain-specific approaches \cite{SWXR19} improved tasks like clinical concept extraction. Unifying contextualized span representations \cite{WWLH19} with dynamic entity types \cite{MMT+24} could thus be a promising direction towards flexible schema evolution.

\medskip 

\stitle{Large Language Models:} Pre-trained Language Models (PLMs) like BERT \cite{DCLT19} and RoBERTa \cite{LOG+19} are usually the go-to options for downstream tasks like concept extraction \cite{FZ22}. However, PLMs require task-specific fine-tuning \cite{ZLS23}. They also struggle to adapt to new domains or entity types without additional labeled data \cite{HLS+21}\cite{GY22}. In contrast, Large Language Models excel at these tasks because they can be instruction-tuned with simple prompts without fine-tuning \cite{RWC+19}. Unlike PLMs, which encode static token-level representations, LLMs operate with a larger parameter space (ranging from a few billion to hundreds of billions) and utilize in-context learning \cite{KGR+22b}. This means that by simply providing a few examples or instructions in the input prompt, these models can adjust their behavior to recognize new entity types and handle domain-specific terminology \cite{BMR+20}. In zero- or few-shot situations, even smaller LLMs (under 7B parameters) often outperform PLMs since they can connect loosely related concept hierarchies \cite{CGS23}. This makes information extraction pipelines more generic.

\medskip

\stitle{Retrieval-Augmented Generation (RAG):} RAG-based techniques can improve LLM output for concept and entity recognition by directly integrating external knowledge sources into the retrieval, generation, and augmentation process~\cite{GXG+23, LPP+20, PPF+21}. RAG dynamically retrieves domain-specific information to address hallucinations and outdated knowledge~\cite{ZZY+24}. This grounding in external data makes predictions more accurate and credible~\cite{PPDG24}. It also enables better handling of rare concepts and entities for flexible schema evolution through continuous updates~\cite{GLT+20}. More recently, Retrieval-Augmented LLMs have demonstrated even greater refinement capabilities, ensuring adaptability in dynamic and knowledge-intensive scenarios~\cite{FDN+24}.

\vspace{-6mm}
\section{Open Problems and Future Directions}
\vspace{-2mm}

To sum up, in order to facilitate data integration that benefits from both structured and unstructured sources, we have to resort to a number of techniques from a multitude of domains. KGs facilitate a provision for virtually managing and representing data in the form of a concept map or semantic network that follows a particular schema~\cite{HBC+21}. On the other hand, unstructured data such as text does not follow any specific structure or schema. \emph{Information Extraction}~\cite{Wilks97, MHL20, DDL+24} techniques from \emph{NLP} can be used in this regard to convert the information embedded in text into a structured representation. Thus facilitating the construction and enhancement~\cite{XZX+23, FRE+20} of structured knowledge sources such as KG~\cite{JM25, KI18, PS18, MLR18}. IE typically consists of a sequence of tasks, comprising linguistic pre-processing, co-reference resolution~\cite{ETE17}, named entity recognition or concept extraction~\cite{EN12}, and relation Extraction. More recent practice to such concept extraction utilizes neural \emph{Language Models} based on transformer and attention mechanisms (i.e., BERT, XLNet, RoBERTa, T5, GPT, Llma, UniversalNER)~\cite{WJB+21, HLS+22, MT22, WSC+22, ZZG+24}. In contrast, popular approaches to RE can be categorized as bootstrapping~\cite{AG00, ECD+04, CBK+10}, distant supervision~\cite{MBSJ09, AVM15, YCL+18, SHS+22}, open-IE or unsupervised~\cite{FSE11, APM15}, and supervised~\cite{HLS+22, ZZC+17, HGY+19, SFLK19, TCSW21} methods. All these approaches can be useful to extract and integrate information from unstructured data sources like text and to populate KGs with structured data.

However, most advanced KGC techniques, essentially, follow the same principles as LM-based techniques for ER or RE and suffer from the same problems: \textbf{(i)} the need for a domain-specific large annotated data with rich context, \textbf{(ii)} the need to re-annotate and re-train when the reference schema changes, \textbf{(iii)} bias towards the most frequent entity type, \textbf{(iv)} inconsistent performance and hallucination and \textbf{(v)} resource-hungry training. Unfortunately, these assumptions do not hold in data integration scenario, as available data is predominantly structured with limited context (e.g., tabular data), making it difficult for LMs to work properly. Also, despite organizations having lots of textual data, those are not annotated. In the absence of these requirements, the performance of LMs quickly decreases \cite{BGMS21}. Further, in a data integration system, the integrated schema evolves, which would require re-annotating the corpus and re-training the model. Indeed, adapting these methods in the presence of structured data, which by definition have limited context, is nowadays an open research area \cite{WSC+22, YNYR20, SDH+23}. Alternatively, zero-shot LLMs such as GPT \cite{Openai23}, despite their massive popularity, are unsuited here due to their unpredictable nature \cite{CZZ23}, inconsistent performance \cite{ASM23} and tendency to hallucinate \cite{CFY+23}. LLMs also struggle to maintain recall and precision with large text corpora in the presence of complex concept categories \cite{ZZG+24}. Furthermore, it is not feasible in terms of resources (i.e., compute power and time) to train or fine-tune a large model like GPT on a case-by-case basis \cite{YHD+22}, a situation that often pertains to organizations typically dealing with their own data for integration purposes.

\vspace{-6mm}
\section{Conclusion}
\vspace{-2mm}

In recent years, there have been significant advances in the use of NLP, SW, and KG technologies for data integration and unstructured data processing. These advances have made this field of research timely and relevant, as organizations increasingly need to integrate and analyze structured and unstructured data to gain insights and make data-driven decisions. The use of natural language processing techniques such as named entity recognition and relation extraction has become crucial for extracting structured information from unstructured data sources like text. These techniques enable the processing and analysis of large volumes of text data, which is essential for integrating unstructured data into structured data models. Semantic web technologies provide a structured and formal vocabulary that facilitates the integration of data from different sources. The representation of data in a machine-understandable format allows for automated processing and reasoning over data. KGs are structured representations of knowledge that enable the representation of entities and relationships in a graph-like structure, which can be queried and analyzed using graph algorithms and other analytical techniques. With the help of recent NLP breakthroughs in language modeling via LLMs, KGs can be populated with data extracted from a variety of sources, including unstructured data such as text. This will allow a more complete and accurate integration of the multi-modal data, which can be queried and analyzed more effectively. Thus, benefiting a wide range of applications in sectors such as healthcare, business, and government.

\bibliographystyle{spbasic_unsorted}
\bibliography{Ataur_Chap1}

\begin{thebibliography}{198}
\providecommand{\natexlab}[1]{#1}
\providecommand{\url}[1]{{#1}}
\providecommand{\urlprefix}{URL }
\expandafter\ifx\csname urlstyle\endcsname\relax
  \providecommand{\doi}[1]{DOI~\discretionary{}{}{}#1}\else
  \providecommand{\doi}{DOI~\discretionary{}{}{}\begingroup \urlstyle{rm}\Url}\fi
\providecommand{\eprint}[2][]{\url{#2}}

\bibitem[{Lemahieu et~al.(2018)Lemahieu, vanden Broucke, and Baesens}]{LVB18}
Lemahieu W, vanden Broucke S, Baesens B (2018) Principles of database management: The practical guide to storing, managing and analyzing big and small data. Cambridge University Press

\bibitem[{Anadiotis et~al.(2022)Anadiotis, Balalau, Conceicao, Galhardas, Haddad, Manolescu, Merabti, and You}]{ABC+22}
Anadiotis AC, Balalau O, Conceicao C, Galhardas H, Haddad MY, Manolescu I, Merabti T, You J (2022) Graph integration of structured, semistructured and unstructured data for data journalism. Information Systems 104:101846

\bibitem[{Guizzardi(2006)}]{Guizzardi06}
Guizzardi G (2006) On ontology, ontologies, conceptualizations, modeling languages, and (meta)models. In: Vasilecas O, Eder J, Caplinskas A (eds) Databases and Information Systems {IV} - Selected Papers from the Seventh International Baltic Conference, DB{\&}IS, {IOS} Press, Lithuania, pp 18--39

\bibitem[{Ilyas and Chu(2019)}]{IC19}
Ilyas IF, Chu X (2019) Data Cleaning. {ACM}

\bibitem[{Chen et~al.(2022)Chen, Yang, and Kim}]{CYK22}
Chen S, Yang S, Kim JK (2022) Nonparametric mass imputation for data integration. Journal of survey statistics and methodology 10(1):1--24

\bibitem[{Nath et~al.(2020)Nath, Hose, Pedersen, Romero, and Bhattacharjee}]{NHP+20}
Nath RPD, Hose K, Pedersen TB, Romero O, Bhattacharjee A (2020) {SETLBI:} an integrated platform for semantic business intelligence. In: Seghrouchni AEF, Sukthankar G, Liu T, van Steen M (eds) Companion of The 2020 Web Conference, {ACM} / {IW3C2}, Taiwan, pp 167--171

\bibitem[{Deb~Nath et~al.(2022)Deb~Nath, Romero, Pedersen, and Hose}]{DRPH22}
Deb~Nath RP, Romero O, Pedersen TB, Hose K (2022) High-level etl for semantic data warehouses. Semantic Web 13(1):85--132

\bibitem[{Alqarni and Pardede(2012)}]{AP12}
Alqarni AA, Pardede E (2012) Integration of data warehouse and unstructured business documents. In: Barolli L, Taniar D, Enokido T, Rahayu JW, Takizawa M (eds) 15th International Conference on Network-Based Information Systems, NBiS, {IEEE} Computer Society, Australia, pp 32--37

\bibitem[{Williams and Poulovassilis(2003)}]{WP03}
Williams D, Poulovassilis A (2003) Combining data integration with natural language technology for the semantic web. In: Tablan V, Bontcheva K, Maynard D (eds) Proceedings of the Workshop on Human Language Technology for the Semantic Web and Web Services at ISWC, CEUR, {USA}, pp 35--42

\bibitem[{Chawathe et~al.(1994)Chawathe, Garcia-Molina, Hammer, Ireland, Papakonstantinou, Ullman, and Widom}]{CGH+94}
Chawathe S, Garcia-Molina H, Hammer J, Ireland K, Papakonstantinou Y, Ullman JD, Widom J (1994) The tsimmis project: Integration of heterogeneous information sources. In: of~Japan IPS (ed) Proceedings of the Conference of the Information Processing Society of Japan (IPSJ), Information Processing Society of Japan, Tokyo, Japan, pp 7--18

\bibitem[{Roth and Schwarz(1997)}]{RS97}
Roth MT, Schwarz PM (1997) Don't scrap it, wrap it! {A} wrapper architecture for legacy data sources. In: Jarke M, Carey MJ, Dittrich KR, Lochovsky FH, Loucopoulos P, Jeusfeld MA (eds) Proceedings of 23rd International Conference on Very Large Data Bases, {VLDB}, Morgan Kaufmann, Greece, pp 266--275

\bibitem[{Levy et~al.(1996)Levy, Rajaraman, and Ordille}]{LRO96}
Levy AY, Rajaraman A, Ordille JJ (1996) Querying heterogeneous information sources using source descriptions. In: Vijayaraman TM, Buchmann AP, Mohan C, Sarda NL (eds) Proceedings of 22th International Conference on Very Large Data Bases, {VLDB}, Morgan Kaufmann, India, pp 251--262

\bibitem[{McBrien and Poulovassilis(2003)}]{MP03}
McBrien P, Poulovassilis A (2003) Data integration by bi-directional schema transformation rules. In: Dayal U, Ramamritham K, Vijayaraman TM (eds) Proceedings of the 19th International Conference on Data Engineering, {ICDE}, {IEEE}, India, pp 227--238

\bibitem[{Friedman et~al.(1999)Friedman, Levy, Millstein et~al.}]{FLMO99}
Friedman M, Levy AY, Millstein TD, et~al. (1999) Navigational plans for data integration. AAAI/IAAI 1999:67--73

\bibitem[{Poulovassilis and McBrien(1998)}]{PM98}
Poulovassilis A, McBrien P (1998) A general formal framework for schema transformation. Data \& Knowledge Engineering 28(1):47--71

\bibitem[{Manola et~al.(2004)Manola, Miller, McBride et~al.}]{MMMO04}
Manola F, Miller E, McBride B, et~al. (2004) Rdf primer. W3C recommendation 10(1-107):6

\bibitem[{Augenstein et~al.(2012)Augenstein, Pad{\'{o}}, and Rudolph}]{APR12}
Augenstein I, Pad{\'{o}} S, Rudolph S (2012) Lodifier: Generating linked data from unstructured text. In: Simperl E, Cimiano P, Polleres A, Corcho {\'{O}}, Presutti V (eds) The Semantic Web: Research and Applications - 9th Extended Semantic Web Conference, {ESWC}, Springer, Greece, pp 210--224

\bibitem[{Touma et~al.(2015)Touma, Romero, and Jovanovic}]{TRJ15}
Touma R, Romero O, Jovanovic P (2015) Supporting data integration tasks with semi-automatic ontology construction. In: Song I, Garcia{-}Alvarado C, Ordonez C (eds) Proceedings of the Eighteenth International Workshop on Data Warehousing and OLAP, {DOLAP}, {ACM}, Australia, pp 89--98

\bibitem[{Jovanovic et~al.(2021)Jovanovic, Nadal, Romero, Abell{\'o}, and Bilalli}]{JNR+21}
Jovanovic P, Nadal S, Romero O, Abell{\'o} A, Bilalli B (2021) Quarry: a user-centered big data integration platform. Information Systems Frontiers 23(1):9--33

\bibitem[{Wang(2017)}]{Wang17}
Wang L (2017) Heterogeneous data and big data analytics. Automatic Control and Information Sciences 3(1):8--15

\bibitem[{Golshan et~al.(2017)Golshan, Halevy, Mihaila, and Tan}]{GHMT17}
Golshan B, Halevy AY, Mihaila GA, Tan W (2017) Data integration: After the teenage years. In: Sallinger E, den Bussche JV, Geerts F (eds) Proceedings of the 36th {ACM} Symposium on Principles of Database Systems, {PODS}, {ACM}, {USA}, pp 101--106

\bibitem[{Doan et~al.(2009)Doan, Naughton, Ramakrishnan, Baid, Chai, Chen, Chen, Chu, DeRose, Gao et~al.}]{DNR+09}
Doan A, Naughton JF, Ramakrishnan R, Baid A, Chai X, Chen F, Chen T, Chu E, DeRose P, Gao B, et~al. (2009) Information extraction challenges in managing unstructured data. ACM SIGMOD Record 37(4):14--20

\bibitem[{Adnan et~al.(2020)Adnan, Akbar, Khor, and Ali}]{AAKA20}
Adnan K, Akbar R, Khor SW, Ali ABA (2020) Role and challenges of unstructured big data in healthcare. In: Sharma N, Kumar A, Garg H, Balas VE (eds) Data Management, Analytics and Innovation, Springer Singapore, pp 301--323

\bibitem[{Buitelaar et~al.(2008)Buitelaar, Cimiano, Frank, Hartung, and Racioppa}]{BCF+08}
Buitelaar P, Cimiano P, Frank A, Hartung M, Racioppa S (2008) Ontology-based information extraction and integration from heterogeneous data sources. International Journal of Human-Computer Studies 66(11):759--788

\bibitem[{G{\'o}mez-P{\'e}rez and Manzano-Macho(2004)}]{GM04}
G{\'o}mez-P{\'e}rez A, Manzano-Macho D (2004) An overview of methods and tools for ontology learning from texts. The knowledge engineering review 19(3):187--212

\bibitem[{Gruber(1993)}]{Gruber93}
Gruber TR (1993) A translation approach to portable ontology specifications. Knowledge acquisition 5(2):199--220

\bibitem[{Maedche and Staab(2000)}]{MS00b}
Maedche A, Staab S (2000) Mining ontologies from text. In: Dieng R, Corby O (eds) 12th International Conference on Knowledge Acquisition, Modeling and Management, {EKAW}, Springer, France, pp 189--202

\bibitem[{Martinez-Rodriguez et~al.(2020)Martinez-Rodriguez, Hogan, and Lopez-Arevalo}]{MHL20}
Martinez-Rodriguez JL, Hogan A, Lopez-Arevalo I (2020) Information extraction meets the semantic web: a survey. Semantic Web 11(2):255--335

\bibitem[{Kertkeidkachorn and Ichise(2018)}]{KI18}
Kertkeidkachorn N, Ichise R (2018) An automatic knowledge graph creation framework from natural language text. IEICE TRANSACTIONS on Information and Systems 101(1):90--98

\bibitem[{Pujara and Singh(2018)}]{PS18}
Pujara J, Singh S (2018) Mining knowledge graphs from text. In: Chang Y, Zhai C, Liu Y, Maarek Y (eds) Proceedings of the Eleventh International Conference on Web Search and Data Mining, {WSDM}, {ACM}, {USA}, pp 789--790

\bibitem[{Martinez-Rodriguez et~al.(2018)Martinez-Rodriguez, Lopez-Arevalo, and Rios-Alvarado}]{MLR18}
Martinez-Rodriguez JL, Lopez-Arevalo I, Rios-Alvarado AB (2018) Openie-based approach for knowledge graph construction from text. Expert Systems with Applications 113:339--355

\bibitem[{Hofer et~al.(2024)Hofer, Obraczka, Saeedi, K{\"{o}}pcke, and Rahm}]{HOS+24}
Hofer M, Obraczka D, Saeedi A, K{\"{o}}pcke H, Rahm E (2024) Construction of knowledge graphs: Current state and challenges. Inf 15(8):509

\bibitem[{Buitelaar et~al.(2005)Buitelaar, Cimiano, and Magnini}]{BCM05}
Buitelaar P, Cimiano P, Magnini B (2005) Ontology learning from text: methods, evaluation and applications. IOS press

\bibitem[{Buitelaar and Cimiano(2008)}]{BC08}
Buitelaar P, Cimiano P (2008) Ontology learning and population: bridging the gap between text and knowledge. Ios Press

\bibitem[{Abney(1997)}]{Abney97}
Abney S (1997) Part-of-speech tagging and partial parsing. In: Young S, Bloothooft G (eds) Corpus-Based Methods in Language and Speech Processing, Springer, pp 118--136

\bibitem[{Shamsfard and Barforoush(2004)}]{SB04}
Shamsfard M, Barforoush AA (2004) Learning ontologies from natural language texts. International journal of human-computer studies 60(1):17--63

\bibitem[{Khadir et~al.(2021)Khadir, Aliane, and Guessoum}]{KAG21}
Khadir AC, Aliane H, Guessoum A (2021) Ontology learning: Grand tour and challenges. Computer Science Review 39:100339

\bibitem[{Ruder(2020)}]{Ruder20}
Ruder S (2020) {NLP}-progress: Tracking progress in natural language processing. https://nlpprogress.com

\bibitem[{Miller(1995)}]{Miller95}
Miller GA (1995) Wordnet: a lexical database for english. Communications of the ACM 38(11):39--41

\bibitem[{Vossen(1998)}]{Vossen98}
Vossen P (1998) A multilingual database with lexical semantic networks. Dordrecht: Kluwer Academic Publishers doi 10:978--94

\bibitem[{Wang et~al.(2020)Wang, Wang, and Fujita}]{WWF20}
Wang Y, Wang M, Fujita H (2020) Word sense disambiguation: A comprehensive knowledge exploitation framework. Knowledge-Based Systems 190:105030

\bibitem[{Magnini and Strapparava(2000)}]{MS00a}
Magnini B, Strapparava C (2000) Experiments in word domain disambiguation for parallel texts. In: Ide N, Véronis J (eds) {ACL} Workshop on Word Senses and Multi-linguality, {ACL}, China, pp 27--33

\bibitem[{Buitelaar and Sacaleanu(2002)}]{BS02}
Buitelaar P, Sacaleanu B (2002) Extending synsets with medical terms. In: Vossen P, Fellbaum C (eds) Proceedings of the First International WordNet Conference (GWC 2002), Central Institute of Indian Languages, Mysore, India, pp 107--117

\bibitem[{Turcato et~al.(2000)Turcato, Popowich, Toole, Fass, Nicholson, and Tisher}]{TPT+00}
Turcato D, Popowich F, Toole J, Fass D, Nicholson D, Tisher G (2000) Adapting a synonym database to specific domains. In: Ide N, Véronis J (eds) ACL Workshop on Recent Advances in Natural Language Processing and Information Retrieval, {ACL}, China, pp 1--11

\bibitem[{Lin and Pantel(2001)}]{LP01}
Lin D, Pantel P (2001) Induction of semantic classes from natural language text. In: Lee D, Schkolnick M, Provost FJ, Srikant R (eds) Proceedings of the seventh {ACM} international conference on Knowledge discovery and data mining, {SIGKDD}, {ACL}, {USA}, pp 317--322

\bibitem[{Lin and Pantel(2002)}]{LP02}
Lin D, Pantel P (2002) Concept discovery from text. In: Tseng SC, Chen TE, Liu YF (eds) 19th International Conference on Computational Linguistics, {COLING}, Morgan Kaufmann, Taiwan, pp 577--583

\bibitem[{Landauer and Dumais(1997)}]{LD97}
Landauer TK, Dumais ST (1997) A solution to plato's problem: The latent semantic analysis theory of acquisition, induction, and representation of knowledge. Psychological review 104(2):211

\bibitem[{Blei et~al.(2003)Blei, Ng, and Jordan}]{BNJ03}
Blei DM, Ng AY, Jordan MI (2003) Latent dirichlet allocation. Journal of machine Learning research 3(Jan):993--1022

\bibitem[{Mikolov et~al.(2013)Mikolov, Sutskever, Chen, Corrado, and Dean}]{MSC+13}
Mikolov T, Sutskever I, Chen K, Corrado GS, Dean J (2013) Distributed representations of words and phrases and their compositionality. In: Burges CJC, Bottou L, Ghahramani Z, Weinberger KQ (eds) 27th Annual Conference on Neural Information Processing Systems, Curran Associates, Inc., {USA}, pp 3111--3119

\bibitem[{Bojanowski et~al.(2017)Bojanowski, Grave, Joulin, and Mikolov}]{BGJM17}
Bojanowski P, Grave E, Joulin A, Mikolov T (2017) Enriching word vectors with subword information. Transactions of the {ACL} 5:135--146

\bibitem[{Pennington et~al.(2014)Pennington, Socher, and Manning}]{PSM14}
Pennington J, Socher R, Manning CD (2014) Glove: Global vectors for word representation. In: Moschitti A, Pang B, Daelemans W (eds) Proceedings of the Conference on Empirical Methods in Natural Language Processing, {EMNLP}, {ACL}, Qatar, pp 1532--1543

\bibitem[{Harispe et~al.(2015)Harispe, Ranwez, Janaqi, and Montmain}]{HRJM15}
Harispe S, Ranwez S, Janaqi S, Montmain J (2015) Semantic Similarity from Natural Language and Ontology Analysis. Morgan {\&} Claypool Publishers

\bibitem[{Al-Aswadi et~al.(2020)Al-Aswadi, Chan, and Gan}]{ACG20a}
Al-Aswadi FN, Chan HY, Gan KH (2020) Automatic ontology construction from text: a review from shallow to deep learning trend. Artificial Intelligence Review 53(6):3901--3928

\bibitem[{Wong et~al.(2012)Wong, Liu, and Bennamoun}]{WLB12}
Wong W, Liu W, Bennamoun M (2012) Ontology learning from text: A look back and into the future. ACM Computing Surveys (CSUR) 44(4):1--36

\bibitem[{G{\'{o}}mez{-}Suta et~al.(2020)G{\'{o}}mez{-}Suta, Echeverry{-}Correa, and Mej{\'{\i}}a}]{GEM20}
G{\'{o}}mez{-}Suta M, Echeverry{-}Correa JD, Mej{\'{\i}}a JAS (2020) Semi-automatic extraction and validation of concepts in ontology learning from texts in spanish. In: Chbeir R, Manolopoulos Y, Akerkar R, Mizera{-}Pietraszko J (eds) The 10th International Conference on Web Intelligence, Mining and Semantics, {WIMS} Biarritz, {ACM}, France, pp 7--16

\bibitem[{Exner and Nugues(2012)}]{EN12}
Exner P, Nugues P (2012) Entity extraction: From unstructured text to dbpedia {RDF} triples. In: Rizzo G, Mendes PN, Charton E, Hellmann S, Kalyanpur A (eds) Proceedings of the Web of Linked Entities Workshop, CEUR-WS.org, {USA}, pp 58--69

\bibitem[{El{-}Kilany et~al.(2017)El{-}Kilany, Tazi, and Ezzat}]{ETE17}
El{-}Kilany A, Tazi NE, Ezzat E (2017) Building relation extraction templates via unsupervised learning. In: Desai BC, Hong J, McClatchey R (eds) Proceedings of the 21st International Database Engineering {\&} Applications Symposium, {IDEAS}, {ACM}, United Kingdom, pp 228--234

\bibitem[{Gamallo et~al.(2002)Gamallo, Gonzalez, Agustini, Lopes, and de~Lima}]{GGA+02}
Gamallo P, Gonzalez M, Agustini A, Lopes G, de~Lima VS (2002) Mapping syntactic dependencies onto semantic relations. In: Aussenac-Gilles N, Maedche A (eds) Proceedings of the ECAI, Workshop on Machine Learning and Natural Language Processing for Ontology Engineering, {INRIA}, France, pp 15--22

\bibitem[{Rajagopal et~al.(2013)Rajagopal, Cambria, Olsher, and Kwok}]{RCOK13}
Rajagopal D, Cambria E, Olsher D, Kwok K (2013) A graph-based approach to commonsense concept extraction and semantic similarity detection. In: Carr L, Laender AHF, L{\'{o}}scio BF, King I, Fontoura M, Vrandecic D, Aroyo L, de~Oliveira JPM, Lima F, Wilde E (eds) 22nd International World Wide Web Conference, {WWW}, {ACM}, Brazil, pp 565--570

\bibitem[{Liu et~al.(2008)Liu, Xu, Zhang, Wang, Yu, and Pan}]{LXZ+08}
Liu Q, Xu K, Zhang L, Wang H, Yu Y, Pan Y (2008) Catriple: Extracting triples from wikipedia categories. In: Domingue J, Anutariya C (eds) The Semantic Web, 3rd Asian Semantic Web Conference, {ASWC}, Springer, Thailand, pp 330--344

\bibitem[{Hu et~al.(2022)Hu, Liu, Sun, Wu, Liu, Zhang, and Peng}]{HLS+22}
Hu W, Liu L, Sun Y, Wu Y, Liu Z, Zhang R, Peng T (2022) Nlire: A natural language inference method for relation extraction. Journal of Web Semantics 72:100686

\bibitem[{Milosevic and Thielemann(2022)}]{MT22}
Milosevic N, Thielemann W (2022) Relationship extraction for knowledge graph creation from biomedical literature. CoRR abs/2201.01647:1--12

\bibitem[{Han et~al.(2011)Han, Kamber, and Pei}]{HKP11}
Han J, Kamber M, Pei J (2011) Data warehousing and online analytical processing. In: Gray J (ed) Data Mining: Concepts and Techniques, Morgan Kaufmann, pp 123--157

\bibitem[{Hearst(1992)}]{Hearst92}
Hearst MA (1992) Automatic acquisition of hyponyms from large text corpora. In: Boitet C (ed) 14th International Conference on Computational Linguistics, {COLING}, ACL, France, pp 539--545

\bibitem[{Buitelaar et~al.(2004)Buitelaar, Olejnik, and Sintek}]{BOS04}
Buitelaar P, Olejnik D, Sintek M (2004) A prot{\'{e}}g{\'{e}} plug-in for ontology extraction from text based on linguistic analysis. In: Bussler C, Davies J, Fensel D, Studer R (eds) The Semantic Web: Research and Applications, First European Semantic Web Symposium, {ESWS}, Springer, Greece, pp 31--44

\bibitem[{Harris(1954)}]{Harris54}
Harris ZS (1954) Distributional structure. Word 10(2-3):146--162

\bibitem[{Firth(1957)}]{Firth57}
Firth JR (1957) A synopsis of linguistic theory, 1930--1955. In: Firth JR (ed) Studies in Linguistic Analysis, Basil Blackwell, pp 1--32

\bibitem[{Sahlgren(2008)}]{Sahlgren08}
Sahlgren M (2008) The distributional hypothesis. Italian Journal of Disability Studies 20:33--53

\bibitem[{Fu et~al.(2014)Fu, Guo, Qin, Che, Wang, and Liu}]{FGQ+14}
Fu R, Guo J, Qin B, Che W, Wang H, Liu T (2014) Learning semantic hierarchies via word embeddings. In: Toutanova K, Wu H (eds) Proceedings of the 52nd Annual Meeting of the {ACL}, {ACL}, {ACL}, {USA}, pp 1199--1209

\bibitem[{Nguyen et~al.(2017)Nguyen, K{\"{o}}per, im~Walde, and Vu}]{NKWV17}
Nguyen KA, K{\"{o}}per M, im~Walde SS, Vu NT (2017) Hierarchical embeddings for hypernymy detection and directionality. In: Palmer M, Hwa R, Riedel S (eds) Proceedings of the Conference on Empirical Methods in Natural Language Processing, {EMNLP}, {ACL}, Denmark, pp 233--243

\bibitem[{Alsuhaibani et~al.(2019)Alsuhaibani, Maehara, and Bollegala}]{AMB19}
Alsuhaibani M, Maehara T, Bollegala D (2019) Joint learning of hierarchical word embeddings from a corpus and a taxonomy. In: McCallum A, Augenstein I, Singh S (eds) 1st Conference on Automated Knowledge Base Construction, {AKBC}, OpenReview, {USA}, pp 1--9

\bibitem[{Peters and Westerst{\aa}hl(2013)}]{PW13}
Peters S, Westerst{\aa}hl D (2013) The semantics of possessives. Language 89(4):713--759

\bibitem[{Al{-}Aswadi et~al.(2020)Al{-}Aswadi, Chan, and Gan}]{ACG20b}
Al{-}Aswadi FN, Chan HY, Gan KH (2020) Extracting semantic concepts and relations from scientific publications by using deep learning. In: Saeed F, Mohammed F, Al{-}Nahari A (eds) Innovative Systems for Intelligent Health Informatics - Data Science, Health Informatics, Intelligent Systems, Smart Computing, {IRICT}, Springer, Malaysia, pp 374--383

\bibitem[{Alobaidi et~al.(2018)Alobaidi, Malik, and Sabra}]{AMS18}
Alobaidi M, Malik KM, Sabra S (2018) Linked open data-based framework for automatic biomedical ontology generation. BMC bioinformatics 19(1):1--13

\bibitem[{Akter and Rahman(2019)}]{AR19}
Akter YA, Rahman MA (2019) Extracting rdf triples from raw text. In: Reza AW, Shahnaz C (eds) 2019 1st International Conference on Advances in Science, Engineering and Robotics Technology (ICASERT), {IEEE}, Bangladesh, pp 1--4

\bibitem[{Han et~al.(2019)Han, Gao, Yao, Ye, Liu, and Sun}]{HGY+19}
Han X, Gao T, Yao Y, Ye D, Liu Z, Sun M (2019) Opennre: An open and extensible toolkit for neural relation extraction. In: Pad{\'{o}} S, Huang R (eds) Proceedings of the Conference on Empirical Methods in Natural Language Processing and the 9th International Joint Conference on Natural Language Processing, {EMNLP-IJCNLP}, {ACL}, China, pp 169--174

\bibitem[{Qiu et~al.(2018)Qiu, Chai, Liu, Gu, Li, and Tian}]{QCL+18}
Qiu J, Chai Y, Liu Y, Gu Z, Li S, Tian Z (2018) Automatic non-taxonomic relation extraction from big data in smart city. IEEE Access 6:74854--74864

\bibitem[{Kaushik and Chatterjee(2018)}]{KC18}
Kaushik N, Chatterjee N (2018) Automatic relationship extraction from agricultural text for ontology construction. Information processing in agriculture 5(1):60--73

\bibitem[{Yahya et~al.(2014)Yahya, Whang, Gupta, and Halevy}]{YWGH14}
Yahya M, Whang S, Gupta R, Halevy AY (2014) Renoun: Fact extraction for nominal attributes. In: Moschitti A, Pang B, Daelemans W (eds) Proceedings of the Conference on Empirical Methods in Natural Language Processing, {EMNLP}, {ACL}, Qatar, pp 325--335

\bibitem[{Mintz et~al.(2009)Mintz, Bills, Snow, and Jurafsky}]{MBSJ09}
Mintz M, Bills S, Snow R, Jurafsky D (2009) Distant supervision for relation extraction without labeled data. In: Su K, Su J, Wiebe J (eds) Proceedings of the 47th Annual Meeting of the {ACL} and the 4th International Joint Conference on Natural Language Processing of the AFNLP, {ACL}, Singapore, pp 1003--1011

\bibitem[{Madaan et~al.(2016)Madaan, Mittal, Mausam, Ramakrishnan, and Sarawagi}]{MMM+16}
Madaan A, Mittal AR, Mausam, Ramakrishnan G, Sarawagi S (2016) Numerical relation extraction with minimal supervision. In: Schuurmans D, Wellman MP (eds) Proceedings of the Thirtieth Conference on Artificial Intelligence, {AAAI}, {AAAI} Press, {USA}, pp 2764--2771

\bibitem[{Nakashole et~al.(2011)Nakashole, Theobald, and Weikum}]{NTW11}
Nakashole N, Theobald M, Weikum G (2011) Scalable knowledge harvesting with high precision and high recall. In: King I, Nejdl W, Li H (eds) Proceedings of the Forth International Conference on Web Search and Web Data Mining, {WSDM}, {ACM}, China, pp 227--236

\bibitem[{Ji et~al.(2022)Ji, Pan, Cambria, Marttinen, and Yu}]{JPC+22}
Ji S, Pan S, Cambria E, Marttinen P, Yu PS (2022) A survey on knowledge graphs: Representation, acquisition, and applications. {IEEE} Trans Neural Networks Learn Syst 33(2):494--514

\bibitem[{Tiwari et~al.(2021)Tiwari, Al-Aswadi, and Gaurav}]{TAG21}
Tiwari S, Al-Aswadi FN, Gaurav D (2021) Recent trends in knowledge graphs: theory and practice. Soft Computing 25(13):8337--8355

\bibitem[{Liang et~al.(2021)Liang, Stockinger, de~Farias, Anisimova, and Gil}]{LSD+21}
Liang S, Stockinger K, de~Farias TM, Anisimova M, Gil M (2021) Querying knowledge graphs in natural language. Journal of big Data 8(1):1--23

\bibitem[{Clancy et~al.(2019)Clancy, Ilyas, Lin, and Cheriton}]{CILC19}
Clancy R, Ilyas IF, Lin J, Cheriton D (2019) Knowledge graph construction from unstructured text with applications to fact verification and beyond. In: Thorne J, Vlachos A, Cocarascu O, Christodoulopoulos C, Mittal A (eds) Proceedings of the Second Workshop on Fact Extraction and VERification (FEVER), {ACL}, China, pp 3--7

\bibitem[{Mao et~al.(2019)Mao, Yao, Heinrich, Hinz, Weber, Wermter, Liu, and Sun}]{MYH+19}
Mao J, Yao Y, Heinrich S, Hinz T, Weber C, Wermter S, Liu Z, Sun M (2019) Bootstrapping knowledge graphs from images and text. Frontiers in neurorobotics 13:93

\bibitem[{Smith et~al.(2022)Smith, Papadopoulos, Braschler, and Stockinger}]{SPBS22}
Smith E, Papadopoulos D, Braschler M, Stockinger K (2022) Lillie: Information extraction and database integration using linguistics and learning-based algorithms. Information Systems 105:101938

\bibitem[{Melnyk et~al.(2022)Melnyk, Dognin, and Das}]{MDD22}
Melnyk I, Dognin PL, Das P (2022) Knowledge graph generation from text. In: Goldberg Y, Kozareva Z, Zhang Y (eds) Findings of the {ACL}: {EMNLP}, {ACL}, {UAE}, pp 1610--1622

\bibitem[{Rossiello et~al.(2023)Rossiello, Chowdhury, Mihindukulasooriya, Cornec, and Gliozzo}]{RCM+23}
Rossiello G, Chowdhury MFM, Mihindukulasooriya N, Cornec O, Gliozzo AM (2023) Knowgl: Knowledge generation and linking from text. In: Williams B, Chen Y, Neville J (eds) Thirty-Seventh {AAAI} Conference on Artificial Intelligence, {AAAI}, {AAAI} Press, {USA}, pp 16476--16478

\bibitem[{Hogan et~al.(2021)Hogan, Blomqvist, Cochez, d'Amato, de~Melo, Guti\'errez, Kirrane, Labra~Gayo, Navigli, Neumaier, Ngonga~Ngomo, Polleres, Rashid, Rula, Schmelzeisen, Sequeda, Staab, and Zimmermann}]{HBC+21}
Hogan A, Blomqvist E, Cochez M, d'Amato C, de~Melo G, Guti\'errez C, Kirrane S, Labra~Gayo JE, Navigli R, Neumaier S, Ngonga~Ngomo AC, Polleres A, Rashid SM, Rula A, Schmelzeisen L, Sequeda JF, Staab S, Zimmermann A (2021) {K}nowledge {G}raphs. Morgan \& Claypool

\bibitem[{Browarnik and Maimon(2015)}]{BM15}
Browarnik A, Maimon O (2015) Ontology learning from text: why the ontology learning layer cake is not viable. International Journal of Signs and Semiotic Systems (IJSSS) 4(2):1--14

\bibitem[{Gillani~Andleeb(2015)}]{GS15}
Gillani~Andleeb S (2015) From text mining to knowledge mining: An integrated framework of concept extraction and categorization for domain ontology. PhD thesis, Budapesti Corvinus Egyetem

\bibitem[{Fleischhacker and V{\"{o}}lker(2011)}]{FV11}
Fleischhacker D, V{\"{o}}lker J (2011) Inductive learning of disjointness axioms. In: Meersman R, Dillon TS, Herrero P, Kumar A, Reichert M, Qing L, Ooi BC, Damiani E, Schmidt DC, White J, Hauswirth M, Hitzler P, Mohania MK (eds) On the Move to Meaningful Internet Systems: {OTM} - Confederated International Conferences: CoopIS, DOA-SVI, and {ODBASE}, Proceedings, Part {II}, Springer, Greece, pp 680--697

\bibitem[{Klarman and Britz(2015)}]{KB15}
Klarman S, Britz K (2015) Towards unsupervised ontology learning from data. In: Booth R, Casini G, Klarman S, Richard G, Varzinczak I (eds) Proceedings of the International Workshop on Defeasible and Ampliative Reasoning, DARe, CEUR, Argentina, pp 33--42

\bibitem[{S{\'a}nchez et~al.(2012)S{\'a}nchez, Moreno, and Del Vasto-Terrientes}]{SMD12}
S{\'a}nchez D, Moreno A, Del Vasto-Terrientes L (2012) Learning relation axioms from text: An automatic web-based approach. Expert Systems with Applications 39(5):5792--5805

\bibitem[{Lenzerini(2002)}]{Lenzerini02}
Lenzerini M (2002) Data integration: {A} theoretical perspective. In: Popa L, Abiteboul S, Kolaitis PG (eds) Proceedings of the Twenty-first {ACM} Symposium on Principles of Database Systems, {PODS}, {ACM}, {USA}, pp 233--246

\bibitem[{Lacroix and Pirotte(1976)}]{LP76}
Lacroix M, Pirotte A (1976) Generalized joins. {SIGMOD} Rec 8(3):14--15

\bibitem[{Rajaraman and Ullman(1996)}]{RU96}
Rajaraman A, Ullman JD (1996) Integrating information by outerjoins and full disjunctions. In: Hull R (ed) Proceedings of the Fifteenth Symposium on Principles of Database Systems, {SIGACT-SIGMOD-SIGART}, {ACM}, Canada, pp 238--248

\bibitem[{Stonebraker and Ilyas(2018)}]{SI18}
Stonebraker M, Ilyas IF (2018) Data integration: The current status and the way forward. {IEEE} Data Eng Bull 41(2):3--9

\bibitem[{Allison et~al.(2006)Allison, Guthrie, and Guthrie}]{AGG06}
Allison B, Guthrie D, Guthrie L (2006) Another look at the data sparsity problem. In: Sojka P, Kopecek I, Pala K (eds) Text, Speech and Dialogue, 9th International Conference, {TSD}, Springer, Czech Republic, pp 327--334

\bibitem[{Xue et~al.(2015)Xue, Qi, Xie, Zhang, Huang, and Li}]{XQX+15}
Xue AY, Qi J, Xie X, Zhang R, Huang J, Li Y (2015) Solving the data sparsity problem in destination prediction. {VLDB} J 24(2):219--243

\bibitem[{Geerts et~al.(2020)Geerts, Mecca, Papotti, and Santoro}]{GMPS20}
Geerts F, Mecca G, Papotti P, Santoro D (2020) Cleaning data with llunatic. {VLDB} 29(4):867--892

\bibitem[{Dallachiesa et~al.(2013)Dallachiesa, Ebaid, Eldawy, Elmagarmid, Ilyas, Ouzzani, and Tang}]{DEE+13}
Dallachiesa M, Ebaid A, Eldawy A, Elmagarmid AK, Ilyas IF, Ouzzani M, Tang N (2013) {NADEEF:} a commodity data cleaning system. In: Ross KA, Srivastava D, Papadias D (eds) Proceedings of the {ACM} International Conference on Management of Data, {SIGMOD}, {ACM}, {USA}, pp 541--552

\bibitem[{Rezig et~al.(2021)Rezig, Ouzzani, Aref, Elmagarmid, Mahmood, and Stonebraker}]{ROA+21}
Rezig EK, Ouzzani M, Aref WG, Elmagarmid AK, Mahmood AR, Stonebraker M (2021) Horizon: Scalable dependency-driven data cleaning. Proc {VLDB} Endow 14(11):2546--2554

\bibitem[{Fan(2015)}]{Fan15}
Fan W (2015) Data quality: From theory to practice. {SIGMOD} Rec 44(3):7--18

\bibitem[{Rekatsinas et~al.(2017)Rekatsinas, Chu, Ilyas, and R{\'{e}}}]{RCIR17}
Rekatsinas T, Chu X, Ilyas IF, R{\'{e}} C (2017) Holoclean: Holistic data repairs with probabilistic inference. Proc {VLDB} Endow 10(11):1190--1201

\bibitem[{Mahdavi and Abedjan(2020)}]{MA20}
Mahdavi M, Abedjan Z (2020) Baran: Effective error correction via a unified context representation and transfer learning. Proc {VLDB} Endow 13(11):1948--1961

\bibitem[{Peng et~al.(2022)Peng, Shen, Tang, Liu, Kou, Nie, Cui, and Yu}]{PST+22}
Peng J, Shen D, Tang N, Liu T, Kou Y, Nie T, Cui H, Yu G (2022) Self-supervised and interpretable data cleaning with sequence generative adversarial networks. Proc {VLDB} Endow 16(3):433--446

\bibitem[{Chu et~al.(2016)Chu, Ilyas, Krishnan, and Wang}]{CIKW16}
Chu X, Ilyas IF, Krishnan S, Wang J (2016) Data cleaning: Overview and emerging challenges. In: {\"{O}}zcan F, Koutrika G, Madden S (eds) Proceedings of the 2016 International Conference on Management of Data, {SIGMOD}, {ACM}, USA, pp 2201--2206

\bibitem[{Chu et~al.(2015)Chu, Morcos, Ilyas, Ouzzani, Papotti, Tang, and Ye}]{CMI+15}
Chu X, Morcos J, Ilyas IF, Ouzzani M, Papotti P, Tang N, Ye Y (2015) {KATARA:} {A} data cleaning system powered by knowledge bases and crowdsourcing. In: Sellis TK, Davidson SB, Ives ZG (eds) Proceedings of the 2015 {ACM} International Conference on Management of Data, {SIGMOD}, {ACM}, Australia, pp 1247--1261

\bibitem[{Pereira et~al.(2024)Pereira, Fonseca, Lopes, and Galhardas}]{PFLG24}
Pereira JLM, Fonseca MJ, Lopes A, Galhardas H (2024) Cleenex: Support for user involvement during an iterative data cleaning process. {ACM} J Data Inf Qual 16(1):6:1--6:26

\bibitem[{Fagin et~al.(2016)Fagin, Kimelfeld, Reiss, and Vansummeren}]{FKRV16}
Fagin R, Kimelfeld B, Reiss F, Vansummeren S (2016) A relational framework for information extraction. ACM SIGMOD Record 44(4):5--16

\bibitem[{Fagin et~al.(2015)Fagin, Kimelfeld, Reiss, and Vansummeren}]{FKRV15}
Fagin R, Kimelfeld B, Reiss F, Vansummeren S (2015) Document spanners: {A} formal approach to information extraction. J {ACM} 62(2):12:1--12:51

\bibitem[{Sarawagi(2008)}]{Sarawagi08}
Sarawagi S (2008) Information extraction. Found Trends Databases 1(3):261--377

\bibitem[{Schneider et~al.(2022)Schneider, Zavala, Mart{\'{\i}}nez, Moro, and Paraiso}]{SZM+22}
Schneider ETR, Zavala RMR, Mart{\'{\i}}nez P, Moro C, Paraiso EC (2022) {UC3M-PUCPR} at semeval-2022 task 11: An ensemble method of transformer-based models for complex named entity recognition. In: Emerson G, Schluter N, Stanovsky G, Kumar R, Palmer A, Schneider N, Singh S, Ratan S (eds) Proceedings of the 16th International Workshop on Semantic Evaluation, SemEval NAACL, {ACL}, {USA}, pp 1448--1456

\bibitem[{Keymanesh et~al.(2022)Keymanesh, Benton, and Dredze}]{KBD22}
Keymanesh M, Benton A, Dredze M (2022) What makes data-to-text generation hard for pretrained language models? CoRR abs/2205.11505:1--16

\bibitem[{Chen et~al.(2023)Chen, Zaharia, and Zou}]{CZZ23}
Chen L, Zaharia M, Zou J (2023) How is chatgpt's behavior changing over time? CoRR abs/2307.09009:1--12

\bibitem[{Arora et~al.(2023)Arora, Singh, and Mausam}]{ASM23}
Arora D, Singh HG, Mausam (2023) Have llms advanced enough? {A} challenging problem solving benchmark for large language models. In: Bouamor H, Pino J, Bali K (eds) Proceedings of the 2023 Conference on Empirical Methods in Natural Language Processing, {EMNLP}, {ACL}, Singapore, pp 7527--7543

\bibitem[{Zhou et~al.(2024)Zhou, Zhang, Gu, Chen, and Poon}]{ZZG+24}
Zhou W, Zhang S, Gu Y, Chen M, Poon H (2024) Universalner: Targeted distillation from large language models for open named entity recognition. In: Kim B, Yue Y, Chaudhuri S, Fragkiadaki K, Khan ME, Sun Y (eds) The Twelfth International Conference on Learning Representations, {ICLR}, OpenReview, Austria, pp 1--12

\bibitem[{Yuan et~al.(2022)Yuan, He, Davis, Zhang, Dao, Chen, Liang, Re, and Zhang}]{YHD+22}
Yuan B, He Y, Davis J, Zhang T, Dao T, Chen B, Liang PS, Re C, Zhang C (2022) Decentralized training of foundation models in heterogeneous environments. Advances in Neural Information Processing Systems 35:25464--25477

\bibitem[{Yang and Kim(2020)}]{YK20}
Yang S, Kim JK (2020) Statistical data integration in survey sampling: A review. Japanese Journal of Statistics and Data Science 3:625--650

\bibitem[{Allen and Cervo(2015)}]{AC15}
Allen M, Cervo D (2015) Data quality management. In: Allen M, Cervo D (eds) Multi-Domain Master Data Management: Advanced MDM and Data Governance in Practice, Morgan Kaufmann, pp 131--160

\bibitem[{Petroni et~al.(2021)Petroni, Piktus, Fan, Lewis, Yazdani, Cao, Thorne, Jernite, Karpukhin, Maillard, Plachouras, Rockt{\"{a}}schel, and Riedel}]{PPF+21}
Petroni F, Piktus A, Fan A, Lewis PSH, Yazdani M, Cao ND, Thorne J, Jernite Y, Karpukhin V, Maillard J, Plachouras V, Rockt{\"{a}}schel T, Riedel S (2021) {KILT:} a benchmark for knowledge intensive language tasks. In: Toutanova K, Rumshisky A, Zettlemoyer L, Hakkani{-}T{\"{u}}r D, Beltagy I, Bethard S, Cotterell R, Chakraborty T, Zhou Y (eds) Proceedings of the Conference of the North American Chapter of the {ACL}: Human Language Technologies, {NAACL-HLT}, {ACL}, Online, pp 2523--2544

\bibitem[{Glass et~al.(2021)Glass, Rossiello, and Gliozzo}]{GRG21}
Glass MR, Rossiello G, Gliozzo A (2021) Zero-shot slot filling with {DPR} and {RAG}. CoRR abs/2104.08610:1--12

\bibitem[{Pigott(2001)}]{Pigott01}
Pigott TD (2001) A review of methods for missing data. Educational research and evaluation 7(4):353--383

\bibitem[{Khan and Hoque(2020)}]{KH20}
Khan SI, Hoque ASML (2020) Sice: an improved missing data imputation technique. Journal of big Data 7(1):1--21

\bibitem[{J{\"a}ger et~al.(2021)J{\"a}ger, Allhorn, and Bie{\ss}mann}]{JAB21}
J{\"a}ger S, Allhorn A, Bie{\ss}mann F (2021) A benchmark for data imputation methods. Frontiers in big Data 4:693674

\bibitem[{Jadhav et~al.(2019)Jadhav, Pramod, and Ramanathan}]{JPR19}
Jadhav A, Pramod D, Ramanathan K (2019) Comparison of performance of data imputation methods for numeric dataset. Applied Artificial Intelligence 33(10):913--933

\bibitem[{Bruni et~al.(2021)Bruni, Daraio, and Aureli}]{BDA21}
Bruni R, Daraio C, Aureli D (2021) Imputation techniques for the reconstruction of missing interconnected data from higher educational institutions. Knowledge-Based Systems 212:106512

\bibitem[{Bie{\ss}mann et~al.(2018)Bie{\ss}mann, Salinas, Schelter, Schmidt, and Lange}]{BSS+18}
Bie{\ss}mann F, Salinas D, Schelter S, Schmidt P, Lange D (2018) "deep" learning for missing value imputationin tables with non-numerical data. In: Cuzzocrea A, Allan J, Paton NW, Srivastava D, Agrawal R, Broder AZ, Zaki MJ, Candan KS, Labrinidis A, Schuster A, Wang H (eds) Proceedings of the 27th {ACM} International Conference on Information and Knowledge Management, {CIKM}, {ACM}, Italy, pp 2017--2025

\bibitem[{Ren et~al.(2023)Ren, Wang, Seklouli, Zhang, and Bouras}]{RWS+23}
Ren L, Wang T, Seklouli AS, Zhang H, Bouras A (2023) A review on missing values for main challenges and methods. Inf Syst 119:102268

\bibitem[{Romero and Salmer{\'o}n(2004)}]{RS04}
Romero V, Salmer{\'o}n A (2004) Multivariate imputation of qualitative missing data using bayesian networks. In: L{\'o}pez-D{\'i}az M, Gil MA, Grzegorzewski P, Hryniewicz O, Lawry J (eds) Soft Methodology and Random Information Systems, Springer Berlin Heidelberg, pp 605--612

\bibitem[{Fries et~al.(2017)Fries, Wu, Ratner, and R{\'{e}}}]{FWRR17}
Fries JA, Wu S, Ratner A, R{\'{e}} C (2017) Swellshark: {A} generative model for biomedical named entity recognition without labeled data. CoRR abs/1704.06360:1--12

\bibitem[{Quimbaya et~al.(2016)Quimbaya, M{\'u}nera, Rivera, Rodr{\'\i}guez, Velandia, Pe{\~n}a, and Labb{\'e}}]{QMR+16}
Quimbaya AP, M{\'u}nera AS, Rivera RAG, Rodr{\'\i}guez JCD, Velandia OMM, Pe{\~n}a AAG, Labb{\'e} C (2016) Named entity recognition over electronic health records through a combined dictionary-based approach. Procedia Computer Science 100:55--61

\bibitem[{Zhao(2004)}]{Zhao04}
Zhao S (2004) Named entity recognition in biomedical texts using an {HMM} model. In: Collier N, Ruch P, Nazarenko A (eds) Proceedings of the International Joint Workshop on Natural Language Processing in Biomedicine and its Applications, NLPBA/BioNLP, COLING, Switzerland, pp 87--90

\bibitem[{Li et~al.(2008)Li, Savova, and Schuler}]{LSS08}
Li D, Savova G, Schuler KK (2008) Conditional random fields and support vector machines for disorder named entity recognition in clinical texts. In: Demner{-}Fushman D, Ananiadou S, Cohen KB, Pestian J, Tsujii J, Webber BL (eds) Proceedings of the Workshop on Current Trends in Biomedical Natural Language Processing, BioNLP, {ACL}, {USA}, pp 94--95

\bibitem[{Rockt{\"a}schel et~al.(2012)Rockt{\"a}schel, Weidlich, and Leser}]{RWL12}
Rockt{\"a}schel T, Weidlich M, Leser U (2012) Chemspot: a hybrid system for chemical named entity recognition. Bioinformatics 28(12):1633--1640

\bibitem[{Leaman et~al.(2015)Leaman, Wei, and Lu}]{LWL15}
Leaman R, Wei CH, Lu Z (2015) tmchem: a high performance approach for chemical named entity recognition and normalization. Journal of cheminformatics 7(1):1--10

\bibitem[{Lenci and Sahlgren(2023)}]{LS23}
Lenci A, Sahlgren M (2023) Distributional semantics. Cambridge University Press

\bibitem[{Devlin et~al.(2019)Devlin, Chang, Lee, and Toutanova}]{DCLT19}
Devlin J, Chang M, Lee K, Toutanova K (2019) {BERT:} pre-training of deep bidirectional transformers for language understanding. In: Burstein J, Doran C, Solorio T (eds) Proceedings of the 2019 Conference of the North American Chapter of the {ACL}: Human Language Technologies, {NAACL-HLT}, {ACL}, {USA}, pp 4171--4186

\bibitem[{Liu et~al.(2019)Liu, Ott, Goyal, Du, Joshi, Chen, Levy, Lewis, Zettlemoyer, and Stoyanov}]{LOG+19}
Liu Y, Ott M, Goyal N, Du J, Joshi M, Chen D, Levy O, Lewis M, Zettlemoyer L, Stoyanov V (2019) Roberta: {A} robustly optimized {BERT} pretraining approach. CoRR abs/1907.11692:1--12

\bibitem[{Alsentzer et~al.(2019)Alsentzer, Murphy, Boag, Weng, Jin, Naumann, and McDermott}]{AMB+19}
Alsentzer E, Murphy JR, Boag W, Weng W, Jin D, Naumann T, McDermott MBA (2019) Publicly available clinical {BERT} embeddings. CoRR abs/1904.03323:1--7

\bibitem[{Bhowmick et~al.(2023)Bhowmick, Dragut, and Meng}]{BDM23}
Bhowmick SS, Dragut EC, Meng W (2023) Globally aware contextual embeddings for named entity recognition in social media streams. In: Li C, Chen L, Manegold S (eds) 39th {IEEE} International Conference on Data Engineering, {ICDE}, {IEEE}, {USA}, pp 1544--1557

\bibitem[{Wang et~al.(2021)Wang, Jiang, Bach, Wang, Huang, Huang, and Tu}]{WJB+21}
Wang X, Jiang Y, Bach N, Wang T, Huang Z, Huang F, Tu K (2021) Automated concatenation of embeddings for structured prediction. In: Zong C, Xia F, Li W, Navigli R (eds) Proceedings of the 59th Annual Meeting of the {ACL} and the 11th International Joint Conference on Natural Language Processing, {ACL/IJCNLP}, {ACL}, Virtual, pp 2643--2660

\bibitem[{Wang et~al.(2022)Wang, Shen, Cai, Wang, Wang, Xie, Huang, Lu, Zhuang, Tu, Lu, and Jiang}]{WSC+22}
Wang X, Shen Y, Cai J, Wang T, Wang X, Xie P, Huang F, Lu W, Zhuang Y, Tu K, Lu W, Jiang Y (2022) {DAMO-NLP} at semeval-2022 task 11: {A} knowledge-based system for multilingual named entity recognition. In: Emerson G, Schluter N, Stanovsky G, Kumar R, Palmer A, Schneider N, Singh S, Ratan S (eds) Proceedings of the 16th International Workshop on Semantic Evaluation, SemEval@NAACL, {ACL}, {USA}, pp 1457--1468

\bibitem[{Lin et~al.(2020)Lin, Li, Xin, Li, and Chen}]{LLX+20}
Lin X, Li H, Xin H, Li Z, Chen L (2020) Kbpearl: A knowledge base population system supported by joint entity and relation linking. Proc VLDB Endow 13(7):1035–1049

\bibitem[{Ji and Grishman(2011)}]{JG11}
Ji H, Grishman R (2011) Knowledge base population: Successful approaches and challenges. In: Lin D, Matsumoto Y, Mihalcea R (eds) The 49th Annual Meeting of the {ACL}: Human Language Technologies, {ACL}, {USA}, pp 1148--1158

\bibitem[{Vannur et~al.(2021)Vannur, Ganesan, Nagalapatti, Patel, and Tippeswamy}]{VGN+21}
Vannur LS, Ganesan B, Nagalapatti L, Patel H, Tippeswamy MN (2021) Data augmentation for fairness in personal knowledge base population. In: Gupta M, Ramakrishnan G (eds) Trends and Applications in Knowledge Discovery and Data Mining, Springer, India, pp 143--152

\bibitem[{Sui et~al.(2021)Sui, Wang, Chen, Liu, Zhao, and Bi}]{SWC+21}
Sui D, Wang C, Chen Y, Liu K, Zhao J, Bi W (2021) Set generation networks for end-to-end knowledge base population. In: Moens M, Huang X, Specia L, Yih SW (eds) Proceedings of the Conference on Empirical Methods in Natural Language Processing, {EMNLP}, {ACL}, Dominican Republic, pp 9650--9660

\bibitem[{Xu et~al.(2023)Xu, Zhou, Xu, Xia, Liu, Chen, and Dou}]{XZX+23}
Xu D, Zhou J, Xu T, Xia Y, Liu J, Chen E, Dou D (2023) Multimodal biological knowledge graph completion via triple co-attention mechanism. In: Li C, Chen L, Manegold S (eds) 39th {IEEE} International Conference on Data Engineering, {ICDE}, {IEEE}, {USA}, pp 3928--3941

\bibitem[{Trajanoska et~al.(2023)Trajanoska, Stojanov, and Trajanov}]{TST23}
Trajanoska M, Stojanov R, Trajanov D (2023) Enhancing knowledge graph construction using large language models. CoRR abs/2305.04676:1--12

\bibitem[{Rao et~al.(2013)Rao, McNamee, and Dredze}]{RMD13}
Rao D, McNamee P, Dredze M (2013) Entity linking: Finding extracted entities in a knowledge base. In: Poibeau T, Saggion H, Piskorski J, Yangarber R (eds) Multi-source, Multilingual Information Extraction and Summarization, Springer, pp 93--115

\bibitem[{Angeli et~al.(2015)Angeli, Premkumar, and Manning}]{APM15}
Angeli G, Premkumar MJJ, Manning CD (2015) Leveraging linguistic structure for open domain information extraction. In: Zong C, Strube M (eds) Proceedings of the 53rd Annual Meeting of the {ACL} and the 7th International Joint Conference on Natural Language Processing of the Asian Federation of Natural Language Processing, {ACL}, Volume 1: Long Papers, The Association for Computer Linguistics, China, pp 344--354

\bibitem[{Amer-Yahia et~al.(2022)Amer-Yahia, Koutrika, Braschler, Calvanese, Lanti, L{\"u}cke-Tieke, Mosca, Mendes~de Farias, Papadopoulos, Patil et~al.}]{AKB+22}
Amer-Yahia S, Koutrika G, Braschler M, Calvanese D, Lanti D, L{\"u}cke-Tieke H, Mosca A, Mendes~de Farias T, Papadopoulos D, Patil Y, et~al. (2022) Inode: building an end-to-end data exploration system in practice. ACM SIGMOD Record 50(4):23--29

\bibitem[{Wilks(1997)}]{Wilks97}
Wilks Y (1997) Information extraction as a core language technology. In: Pazienza MT (ed) Information Extraction: {A} Multidisciplinary Approach to an Emerging Information Technology, International Summer School, SCIE-97, Springer, Italy, pp 1--9

\bibitem[{Dagdelen et~al.(2024)Dagdelen, Dunn, Lee, Walker, Rosen, Ceder, Persson, and Jain}]{DDL+24}
Dagdelen J, Dunn A, Lee S, Walker N, Rosen AS, Ceder G, Persson KA, Jain A (2024) Structured information extraction from scientific text with large language models. Nature Communications 15(1):1418

\bibitem[{Fern{\`{a}}ndez{-}Ca{\~{n}}ellas et~al.(2020)Fern{\`{a}}ndez{-}Ca{\~{n}}ellas, Rimmek, Espadaler, Garolera, Barja, Codina, Sastre, Gir{\'{o}}{-}i{-}Nieto, Riveiro, and Bou{-}Balust}]{FRE+20}
Fern{\`{a}}ndez{-}Ca{\~{n}}ellas D, Rimmek JM, Espadaler J, Garolera B, Barja A, Codina M, Sastre M, Gir{\'{o}}{-}i{-}Nieto X, Riveiro JC, Bou{-}Balust E (2020) Enhancing online knowledge graph population with semantic knowledge. In: Pan JZ, Tamma V, d'Amato C, Janowicz K, Fu B, Polleres A, Seneviratne O, Kagal L (eds) 19th International Semantic Web Conference, {ISWC}, Springer, Greece, pp 183--200

\bibitem[{Jurafsky and Martin(2025)}]{JM25}
Jurafsky D, Martin JH (2025) Speech and language processing: An introduction to natural language processing, computational linguistics, and speech recognition. \url{https://web.stanford.edu/~jurafsky/slp3/}

\bibitem[{Fernandez et~al.(2023)Fernandez, Elmore, Franklin, Krishnan, and Tan}]{FEF+23}
Fernandez RC, Elmore AJ, Franklin MJ, Krishnan S, Tan C (2023) How large language models will disrupt data management. Proceedings of the VLDB Endowment 16(11):3302--3309

\bibitem[{Zhang et~al.(2017)Zhang, Zhong, Chen, Angeli, and Manning}]{ZZC+17}
Zhang Y, Zhong V, Chen D, Angeli G, Manning CD (2017) Position-aware attention and supervised data improve slot filling. In: Palmer M, Hwa R, Riedel S (eds) Proceedings of the Conference on Empirical Methods in Natural Language Processing, {EMNLP}, {ACL}, Denmark, pp 35--45

\bibitem[{Glass et~al.(2021)Glass, Rossiello, Chowdhury, and Gliozzo}]{GRCG21}
Glass MR, Rossiello G, Chowdhury MFM, Gliozzo A (2021) Robust retrieval augmented generation for zero-shot slot filling. In: Moens M, Huang X, Specia L, Yih SW (eds) Proceedings of the 2021 Conference on Empirical Methods in Natural Language Processing, {EMNLP}, ACL, Dominican Republic, pp 1939--1949

\bibitem[{Zhou and Chen(2021)}]{ZC21}
Zhou W, Chen M (2021) Learning from noisy labels for entity-centric information extraction. In: Moens M, Huang X, Specia L, Yih SW (eds) Proceedings of the Conference on Empirical Methods in Natural Language Processing, {EMNLP}, {ACL}, Dominican Republic, pp 5381--5392

\bibitem[{Rahman et~al.(2024)Rahman, Nadal, Romero, and Sacharidis}]{RNRS24}
Rahman MA, Nadal S, Romero O, Sacharidis D (2024) Mitigating data sparsity in integrated data through text conceptualization. In: Das G, Sellis T, He B (eds) 40th {IEEE} International Conference on Data Engineering, {ICDE}, {IEEE}, The Netherlands, pp 3490--3504

\bibitem[{Fel et~al.(2023)Fel, Boutin, B{\'{e}}thune, Cad{\`{e}}ne, Moayeri, And{\'{e}}ol, Chalvidal, and Serre}]{FBB+23}
Fel T, Boutin V, B{\'{e}}thune L, Cad{\`{e}}ne R, Moayeri M, And{\'{e}}ol L, Chalvidal M, Serre T (2023) A holistic approach to unifying automatic concept extraction and concept importance estimation. In: Oh A, Naumann T, Globerson A, Saenko K, Hardt M, Levine S (eds) Advances in Neural Information Processing Systems 36: Annual Conference on Neural Information Processing Systems, NeurIPS, Curran Associates, Inc., {USA}, pp 1--12

\bibitem[{Taill{\'{e}} et~al.(2020)Taill{\'{e}}, Guigue, and Gallinari}]{TGG20}
Taill{\'{e}} B, Guigue V, Gallinari P (2020) Contextualized embeddings in named-entity recognition: An empirical study on generalization. In: Jose JM, Yilmaz E, Magalh{\~{a}}es J, Castells P, Ferro N, Silva MJ, Martins F (eds) Advances in Information Retrieval - 42nd European Conference on {IR} Research, {ECIR}, Springer, Portugal, pp 383--391

\bibitem[{Akbik et~al.(2019)Akbik, Bergmann, and Vollgraf}]{ABV19}
Akbik A, Bergmann T, Vollgraf R (2019) Pooled contextualized embeddings for named entity recognition. In: Burstein J, Doran C, Solorio T (eds) Proceedings of the 2019 Conference of the North American Chapter of the {ACL}: Human Language Technologies, {NAACL-HLT} 2019, Volume 1 (Long and Short Papers), {ACL}, USA, pp 724--728

\bibitem[{Si et~al.(2019)Si, Wang, Xu, and Roberts}]{SWXR19}
Si Y, Wang J, Xu H, Roberts K (2019) Enhancing clinical concept extraction with contextual embeddings. Journal of the American Medical Informatics Association 26(11):1297--1304

\bibitem[{Wadden et~al.(2019)Wadden, Wennberg, Luan, and Hajishirzi}]{WWLH19}
Wadden D, Wennberg U, Luan Y, Hajishirzi H (2019) Entity, relation, and event extraction with contextualized span representations. In: Inui K, Jiang J, Ng V, Wan X (eds) Proceedings of the 2019 Conference on Empirical Methods in Natural Language Processing and the 9th International Joint Conference on Natural Language Processing, {EMNLP-IJCNLP}, {ACL}, China, pp 5783--5788

\bibitem[{Martinelli et~al.(2024)Martinelli, Molfese, Tedeschi, Fern{\'a}ndez~Castro, and Navigli}]{MMT+24}
Martinelli G, Molfese F, Tedeschi S, Fern{\'a}ndez~Castro A, Navigli R (2024) {CNER}: Concept and named entity recognition. In: Duh K, Gomez H, Bethard S (eds) Proceedings of the 2024 Conference of the North American Chapter of the {ACL}: Human Language Technologies, {NAACL}, {ACL}, Mexico, pp 8336--8351

\bibitem[{Fang and Zhang(2022)}]{FZ22}
Fang Y, Zhang Y (2022) Data-efficient concept extraction from pre-trained language models for commonsense explanation generation. In: Goldberg Y, Kozareva Z, Zhang Y (eds) Findings of the {ACL}: {EMNLP}, {ACL}, United Arab Emirates, pp 5883--5893

\bibitem[{Zhang et~al.(2023)Zhang, Liu, and Shao}]{ZLS23}
Zhang Z, Liu B, Shao J (2023) Fine-tuning happens in tiny subspaces: Exploring intrinsic task-specific subspaces of pre-trained language models. In: Rogers A, Boyd{-}Graber JL, Okazaki N (eds) Proceedings of the 61st Annual Meeting of the Association for Computational Linguistics, {ACL}, {ACL}, Canada, pp 1701--1713

\bibitem[{Huang et~al.(2021)Huang, Li, Subudhi, Jose, Balakrishnan, Chen, Peng, Gao, and Han}]{HLS+21}
Huang J, Li C, Subudhi K, Jose D, Balakrishnan S, Chen W, Peng B, Gao J, Han J (2021) Few-shot named entity recognition: An empirical baseline study. In: Moens M, Huang X, Specia L, Yih SW (eds) Proceedings of the Conference on Empirical Methods in Natural Language Processing, {EMNLP}, {ACL}, Dominican Republic, pp 10408--10423

\bibitem[{Guo and Yu(2022)}]{GY22}
Guo X, Yu H (2022) On the domain adaptation and generalization of pretrained language models: {A} survey. CoRR abs/2211.03154:1--12

\bibitem[{Radford et~al.(2019)Radford, Wu, Child, Luan, Amodei, Sutskever et~al.}]{RWC+19}
Radford A, Wu J, Child R, Luan D, Amodei D, Sutskever I, et~al. (2019) Language models are unsupervised multitask learners. OpenAI blog 1(8):9

\bibitem[{Kojima et~al.(2022)Kojima, Gu, Reid, Matsuo, and Iwasawa}]{KGR+22b}
Kojima T, Gu SS, Reid M, Matsuo Y, Iwasawa Y (2022) Large language models are zero-shot reasoners. Advances in neural information processing systems 35:22199--22213

\bibitem[{Brown et~al.(2020)Brown, Mann, Ryder, Subbiah, Kaplan, Dhariwal, Neelakantan, Shyam, Sastry, Askell, Agarwal, Herbert{-}Voss, Krueger, Henighan, Child, Ramesh, Ziegler, Wu, Winter, Hesse, Chen, Sigler, Litwin, Gray, Chess, Clark, Berner, McCandlish, Radford, Sutskever, and Amodei}]{BMR+20}
Brown TB, Mann B, Ryder N, Subbiah M, Kaplan J, Dhariwal P, Neelakantan A, Shyam P, Sastry G, Askell A, Agarwal S, Herbert{-}Voss A, Krueger G, Henighan T, Child R, Ramesh A, Ziegler DM, Wu J, Winter C, Hesse C, Chen M, Sigler E, Litwin M, Gray S, Chess B, Clark J, Berner C, McCandlish S, Radford A, Sutskever I, Amodei D (2020) Language models are few-shot learners. In: Larochelle H, Ranzato M, Hadsell R, Balcan M, Lin H (eds) Advances in Neural Information Processing Systems 33: Annual Conference on Neural Information Processing Systems, NeurIPS, Curran Associates, Inc., Virtual, pp 1877--1901

\bibitem[{Christiansen et~al.(2023)Christiansen, Gammelgaard, and S{\o}gaard}]{CGS23}
Christiansen JG, Gammelgaard M, S{\o}gaard A (2023) Large language models partially converge toward human-like concept organization. In: Sanborn S, Shewmake C, Azeglio S, Miolane N (eds) Proceedings of the 2nd NeurIPS Workshop on Symmetry and Geometry in Neural Representations, PMLR, USA, pp 346--365

\bibitem[{Gao et~al.(2023)Gao, Xiong, Gao, Jia, Pan, Bi, Dai, Sun, Guo, Wang, and Wang}]{GXG+23}
Gao Y, Xiong Y, Gao X, Jia K, Pan J, Bi Y, Dai Y, Sun J, Guo Q, Wang M, Wang H (2023) Retrieval-augmented generation for large language models: {A} survey. CoRR abs/2312.10997:1--12

\bibitem[{Lewis et~al.(2020)Lewis, Perez, Piktus, Petroni, Karpukhin, Goyal, K{\"u}ttler, Lewis, Yih, Rockt{\"a}schel et~al.}]{LPP+20}
Lewis P, Perez E, Piktus A, Petroni F, Karpukhin V, Goyal N, K{\"u}ttler H, Lewis M, Yih Wt, Rockt{\"a}schel T, et~al. (2020) Retrieval-augmented generation for knowledge-intensive nlp tasks. Advances in Neural Information Processing Systems 33:9459--9474

\bibitem[{Zhao et~al.(2024)Zhao, Zhang, Yu, Wang, Geng, Fu, Yang, Zhang, and Cui}]{ZZY+24}
Zhao P, Zhang H, Yu Q, Wang Z, Geng Y, Fu F, Yang L, Zhang W, Cui B (2024) Retrieval-augmented generation for ai-generated content: {A} survey. CoRR abs/2402.19473:1--15

\bibitem[{Posedaru et~al.(2024)Posedaru, Pantelimon, Dulgheru, and Georgescu}]{PPDG24}
Posedaru BS, Pantelimon FV, Dulgheru MN, Georgescu TM (2024) Artificial intelligence text processing using retrieval-augmented generation: Applications in business and education fields. Proceedings of the International Conference on Business Excellence 18:209 -- 222

\bibitem[{Guu et~al.(2020)Guu, Lee, Tung, Pasupat, and Chang}]{GLT+20}
Guu K, Lee K, Tung Z, Pasupat P, Chang M (2020) Retrieval augmented language model pre-training. In: III HD, Singh A (eds) Proceedings of the 37th International Conference on Machine Learning, {ICML}, {PMLR}, Virtual, pp 3929--3938

\bibitem[{Fan et~al.(2024)Fan, Ding, Ning, Wang, Li, Yin, Chua, and Li}]{FDN+24}
Fan W, Ding Y, Ning L, Wang S, Li H, Yin D, Chua T, Li Q (2024) A survey on {RAG} meeting llms: Towards retrieval-augmented large language models. In: Baeza{-}Yates R, Bonchi F (eds) Proceedings of the 30th {ACM} Conference on Knowledge Discovery and Data Mining, {KDD}, {ACM}, Spain, pp 6491--6501

\bibitem[{Agichtein and Gravano(2000)}]{AG00}
Agichtein E, Gravano L (2000) Snowball: extracting relations from large plain-text collections. In: Nürnberg PJ, Hicks DL, Furuta RK (eds) Proceedings of the Fifth {ACM} Conference on Digital Libraries, {ACM}, {USA}, pp 85--94

\bibitem[{Etzioni et~al.(2004)Etzioni, Cafarella, Downey, Kok, Popescu, Shaked, Soderland, Weld, and Yates}]{ECD+04}
Etzioni O, Cafarella MJ, Downey D, Kok S, Popescu A, Shaked T, Soderland S, Weld DS, Yates A (2004) Web-scale information extraction in knowitall: (preliminary results). In: Feldman SI, Uretsky M, Najork M, Wills CE (eds) Proceedings of the 13th international conference on World Wide Web, {WWW}, {ACM}, {USA}, pp 100--110

\bibitem[{Carlson et~al.(2010)Carlson, Betteridge, Kisiel, Settles, Jr., and Mitchell}]{CBK+10}
Carlson A, Betteridge J, Kisiel B, Settles B, Jr ERH, Mitchell TM (2010) Toward an architecture for never-ending language learning. In: Fox M, Poole D (eds) Proceedings of the Twenty-Fourth Conference on Artificial Intelligence, {AAAI}, {AAAI} Press, {USA}, pp 1306--1313

\bibitem[{Augenstein et~al.(2015)Augenstein, Vlachos, and Maynard}]{AVM15}
Augenstein I, Vlachos A, Maynard D (2015) Extracting relations between non-standard entities using distant supervision and imitation learning. In: M{\`{a}}rquez L, Callison{-}Burch C, Su J, Pighin D, Marton Y (eds) Proceedings of the 2015 Conference on Empirical Methods in Natural Language Processing, {EMNLP}, The {ACL}, Portugal, pp 747--757

\bibitem[{Yang et~al.(2018)Yang, Chen, Li, He, and Zhang}]{YCL+18}
Yang Y, Chen W, Li Z, He Z, Zhang M (2018) Distantly supervised {NER} with partial annotation learning and reinforcement learning. In: Bender EM, Derczynski L, Isabelle P (eds) Proceedings of the 27th International Conference on Computational Linguistics, {COLING}, {ACL}, {USA}, pp 2159--2169

\bibitem[{Shang et~al.(2022)Shang, Huang, Sun, Wei, and Mao}]{SHS+22}
Shang YM, Huang H, Sun X, Wei W, Mao XL (2022) A pattern-aware self-attention network for distant supervised relation extraction. Information Sciences 584:269--279

\bibitem[{Fader et~al.(2011)Fader, Soderland, and Etzioni}]{FSE11}
Fader A, Soderland S, Etzioni O (2011) Identifying relations for open information extraction. In: Barzilay R, Johnson M (eds) Proceedings of the 2011 Conference on Empirical Methods in Natural Language Processing, {EMNLP}, {ACL}, {UK}, pp 1535--1545

\bibitem[{Soares et~al.(2019)Soares, FitzGerald, Ling, and Kwiatkowski}]{SFLK19}
Soares LB, FitzGerald N, Ling J, Kwiatkowski T (2019) Matching the blanks: Distributional similarity for relation learning. In: Korhonen A, Traum DR, M{\`{a}}rquez L (eds) Proceedings of the 57th Conference of the {ACL}, {ACL}, {ACL}, Italy, pp 2895--2905

\bibitem[{Tian et~al.(2021)Tian, Chen, Song, and Wan}]{TCSW21}
Tian Y, Chen G, Song Y, Wan X (2021) Dependency-driven relation extraction with attentive graph convolutional networks. In: Zong C, Xia F, Li W, Navigli R (eds) Proceedings of the 59th Annual Meeting of the {ACL} and the 11th International Joint Conference on Natural Language Processing, {ACL/IJCNLP}, {ACL}, Virtual, pp 4458--4471

\bibitem[{Bender et~al.(2021)Bender, Gebru, McMillan{-}Major, and Shmitchell}]{BGMS21}
Bender EM, Gebru T, McMillan{-}Major A, Shmitchell S (2021) On the dangers of stochastic parrots: Can language models be too big? In: Elish MC, Isaac W, Zemel RS (eds) FAccT '21: 2021 {ACM} Conference on Fairness, Accountability, and Transparency, Virtual Event, {ACM}, Canada, pp 610--623

\bibitem[{Yin et~al.(2020)Yin, Neubig, Yih, and Riedel}]{YNYR20}
Yin P, Neubig G, Yih W, Riedel S (2020) Tabert: Pretraining for joint understanding of textual and tabular data. In: Jurafsky D, Chai J, Schluter N, Tetreault JR (eds) Proceedings of the 58th Annual Meeting of the Association for Computational Linguistics, {ACL}, {ACL}, Online, pp 8413--8426

\bibitem[{Sedlakova et~al.(2023)Sedlakova, Daniore, Horn~Wintsch, Wolf, Stanikic, Haag, Sieber, Schneider, Staub, Alois~Ettlin et~al.}]{SDH+23}
Sedlakova J, Daniore P, Horn~Wintsch A, Wolf M, Stanikic M, Haag C, Sieber C, Schneider G, Staub K, Alois~Ettlin D, et~al. (2023) Challenges and best practices for digital unstructured data enrichment in health research: a systematic narrative review. PLOS Digital Health 2(10):e0000347

\bibitem[{OpenAI(2023)}]{Openai23}
OpenAI (2023) {GPT-4} technical report. CoRR abs/2303.08774:1--98

\bibitem[{Chen et~al.(2023)Chen, Fu, Yuan, Wen, Fan, Liu, Zhang, Li, and Xiao}]{CFY+23}
Chen Y, Fu Q, Yuan Y, Wen Z, Fan G, Liu D, Zhang D, Li Z, Xiao Y (2023) Hallucination detection: Robustly discerning reliable answers in large language models. In: Frommholz I, Hopfgartner F, Lee M, Oakes M, Lalmas M, Zhang M, Santos RLT (eds) Proceedings of the 32nd {ACM} International Conference on Information and Knowledge Management, {CIKM}, {ACM}, United Kingdom, pp 245--255

\end{thebibliography}

%

\backmatter
\appendix


\end{document}